\documentclass[10pt,twocolumn,letterpaper]{article}

\usepackage{cvpr}              
\usepackage{wrapfig}
\usepackage{xspace}
\usepackage{graphicx}
\usepackage{amsmath}
\usepackage{amssymb}
\usepackage{float} 

\definecolor{cvprblue}{rgb}{0.21,0.49,0.74}
\usepackage[pagebackref,breaklinks,colorlinks,allcolors=cvprblue]{hyperref}
\usepackage{caption}
\usepackage{cuted}

\def\paperID{5028} 

\def\confName{CVPR}
\def\confYear{2025}

\title{\vspace{-10mm} Navigation World Models}

\author{
\textbf{Amir Bar}\textsuperscript{1} \qquad \textbf{Gaoyue Zhou}\textsuperscript{2} \qquad \textbf{Danny Tran}\textsuperscript{3} \qquad \textbf{Trevor Darrell}\textsuperscript{3} \qquad \textbf{Yann LeCun}\textsuperscript{1,2} \\
\textsuperscript{1}FAIR at Meta \qquad \qquad \textsuperscript{2}New York University \qquad \qquad  \textsuperscript{3}Berkeley AI Research
\vspace{-5mm}
}

\begin{document}

\maketitle
\begin{strip}
\centering
\href{https://www.amirbar.net/nwm/index.html}{
\includegraphics[width=1\textwidth]{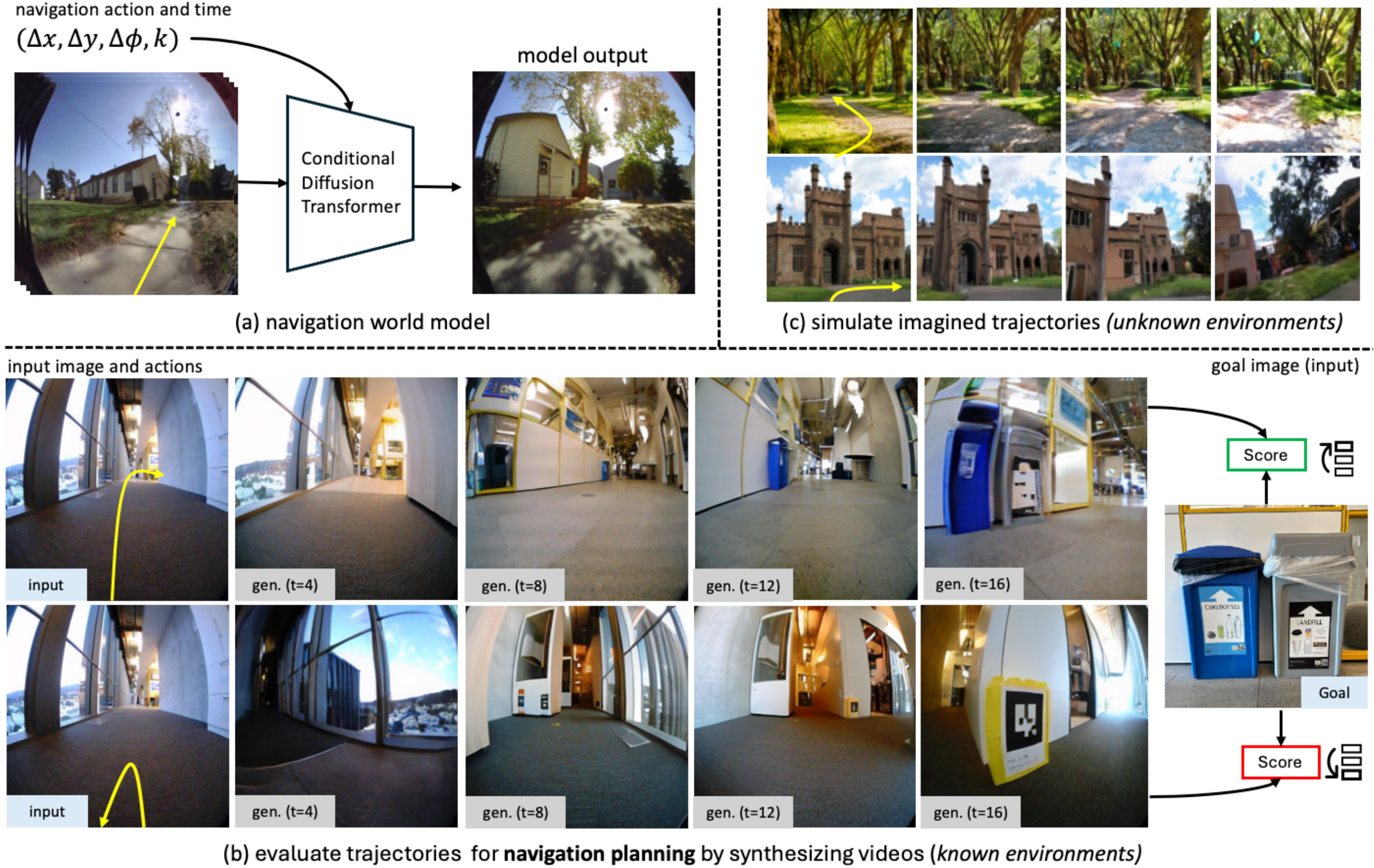}}
\captionsetup{type=figure}
\caption{We train a Navigation World Model (NWM) from video footage of robots and their associated navigation actions (a). After training, NWM can evaluate trajectories by synthesizing their videos and scoring the final frame's similarity with the goal (b). We use NWM to plan from scratch or rank experts navigation trajectories, improving downstream visual navigation performance. In \textit{unknown environments}, NWM can simulate imagined trajectories from a single image (c). In all examples above, the input to the model is the first image and actions, then the model auto-regressively synthesizes future observations. \textbf{Click on the image to view examples in a browser}.}
\label{fig:teaser}
\end{strip}
\begin{abstract}
Navigation is a fundamental skill of agents with visual-motor capabilities. We introduce a Navigation World Model (NWM), a controllable video generation model that predicts future visual observations based on past observations and navigation actions. To capture complex environment dynamics, NWM employs a Conditional Diffusion Transformer (CDiT), trained on a diverse collection of egocentric videos of both human and robotic agents, and scaled up to 1 billion parameters. In familiar environments, NWM can plan navigation trajectories by simulating them and evaluating whether they achieve the desired goal. Unlike supervised navigation policies with fixed behavior, NWM can dynamically incorporate constraints during planning. Experiments demonstrate its effectiveness in planning trajectories from scratch or by ranking trajectories sampled from an external policy. Furthermore, NWM leverages its learned visual priors to imagine trajectories in unfamiliar environments from a single input image, making it a flexible and powerful tool for next-generation navigation systems\footnote{Project page: \url{https://amirbar.net/nwm}}.
\vspace{-13mm}
\end{abstract}    
\section{Introduction}
\label{sec:intro}
Navigation is a fundamental skill for any organism with vision, playing a crucial role in survival by allowing agents to locate food, shelter, and avoid predators. In order to successfully navigate environments, smart agents primarily rely on vision, allowing them to construct representations of their surroundings to assess distances and capture landmarks in the environment, all useful for planning a navigation route.

When human agents plan, they often imagine their future trajectories considering constraints and counterfactuals. On the other hand, current state-of-the-art robotics navigation policies~\citep{sridhar2024nomad, shah2023gnm} are ``hard-coded'', and after training, new constraints cannot be easily introduced (e.g. ``no left turns''). Another limitation of current supervised visual navigation models is that they cannot dynamically allocate more computational resources to address hard problems. We aim to design a new model that can mitigate these issues.

In this work, we propose a Navigation World Model (NWM), trained to predict the future representation of a video frame based on past frame representation(s) and action(s) (see Figure~\ref{fig:teaser}(a)). NWM is trained on video footage and navigation actions collected from various robotic agents. After training, NWM is used to plan novel navigation trajectories by simulating potential navigation plans and verifying if they reach a target goal (see Figure~\ref{fig:teaser}(b)). To evaluate its navigation skills, we test NWM in \emph{known environments}, assessing its ability to plan novel trajectories either independently or by ranking an external navigation policy. In the planning setup, we use NWM in a Model Predictive Control (MPC) framework, optimizing the action sequence that enables NWM to reach a target goal. In the ranking setup, we assume access to an existing navigation policy, such as NoMaD~\citep{sridhar2024nomad}, which allows us to sample trajectories, simulate them using NWM, and select the best ones. Our NWM achieves state-of-the-art standalone performance and competitive results when combined with existing methods.

NWM is conceptually similar to recent diffusion-based world models for offline model-based reinforcement learning, such as DIAMOND~\citep{alonso2024diffusionworldmodelingvisual} and GameNGen~\citep{valevski2024diffusion}. However, unlike these models, NWM is trained across a wide range of environments and embodiments, leveraging the diversity of navigation data from robotic and human agents. This allows us to train a large diffusion transformer model capable of scaling effectively with model size and data to adapt to multiple environments. Our approach also shares similarities with Novel View Synthesis (NVS) methods like NeRF~\citep{mildenhall2021nerf}, Zero-1-2-3~\citep{liu2023zero}, and GDC~\citep{vanhoorick2024gcd}, from which we draw inspiration. However, unlike NVS approaches, our goal is to train a single model for navigation across diverse environments and model temporal dynamics from natural videos, without relying on 3D priors.

To learn a NWM, we propose a novel Conditional Diffusion Transformer (CDiT), trained to predict the next image state given past image states and actions as context. Unlike a DiT~\citep{Peebles_2023_ICCV}, CDiT’s computational complexity is linear with respect to the number of context frames, and it scales favorably for models trained up to $1B$ parameters across diverse environments and embodiments, requiring $4\times$ fewer FLOPs compared to a standard DiT while achieving better future prediction results.

In unknown environments, our results show that NWM benefits from training on unlabeled, action- and reward-free video data from Ego4D. Qualitatively, we observe improved video prediction and generation performance on single images (see Figure~\ref{fig:teaser}(c)). Quantitatively, with additional unlabeled data, NWM produces more accurate predictions when evaluated on the held-out Stanford Go~\citep{hirose2018gonet} dataset.

Our contributions are as follows. We introduce a Navigation World Model (NWM) and propose a novel Conditional Diffusion Transformer (CDiT), which scales efficiently up to $1B$ parameters with significantly reduced computational requirements compared to standard DiT. We train CDiT on video footage and navigation actions from diverse robotic agents, enabling planning by simulating navigation plans independently or alongside external navigation policies, achieving state-of-the-art visual navigation performance. Finally, by training NWM on action- and reward-free video data, such as Ego4D, we demonstrate improved video prediction and generation performance in unseen environments.
\section{Related Work}
Goal conditioned visual navigation is an important task in robotics requiring both perception and planning skills~\citep{sridhar2024nomad,shahvint,pathak2018zero, mirowski2022learning, chaplotlearning, fu2022coupling,frey2023fast}. Given context image(s) and an image specifying the navigation goals, goal-conditioned visual navigation models~\citep{sridhar2024nomad,shahvint} aim to generate a viable path towards the goal if the environment is known, or to explore it otherwise. Recent visual navigation methods like NoMaD~\citep{sridhar2024nomad} train a diffusion policy via behavior cloning and temporal distance objective to follow goals in the conditional setting or to explore new environments in the unconditional setting. Previous approaches like Active Neural SLAM~\citep{chaplotlearning} used neural SLAM together with analytical planners to plan trajectories in the $3D$ environment, while other approaches like~\citep{chenlearning} learn policies via reinforcement learning. Here we show that world models can use exploratory data to plan or improve existing navigation policies.

Differently than in learning a policy, the goal of a world model~\citep{ha2018world} is to simulate the environment, e.g. given the current state and action to predict the next state and an associated reward. Previous works have shown that jointly learning a policy and a world model can improve sample efficiency on Atari~\citep{hafnermastering,hafnerdream,alonso2024diffusionworldmodelingvisual}, simulated robotics environments~\citep{seo2023masked}, and even when applied to real world robots~\citep{wu2023daydreamer}. More recently, \cite{hansentd} proposed to use a single world model that is shared across tasks by introducing action and task embeddings while~\cite{yanglearning, lin2024learningmodelworldlanguage} proposed to describe actions in language, and~\cite{bruce2024genie} proposed to learn latent actions. World models were also explored in the context of game simulation. DIAMOND~\citep{alonso2024diffusionworldmodelingvisual} and GameNGen~\citep{valevski2024diffusion} propose to use diffusion models to learn game engines of computer games like Atari and Doom. Our work is inspired by these works, and we aim to learn a single general diffusion video transformer that can be shared across many environments and different embodiments for navigation.

In computer vision, generating videos has been a long standing challenge~\citep{kondratyukvideopoet,blattmann2023stable,girdhar2023emu,yu2023magvit,ho2022imagen,tulyakov2018mocogan,bar2024lumiere}. Most recently, there has been tremendous progress with text-to-video synthesis with methods like Sora~\citep{brooks2024video} and MovieGen~\citep{polyak2024movie}. Past works proposed to control video synthesis given structured action-object class categories~\citep{Tulyakov:2018:MoCoGAN} or Action Graphs~\citep{bar2021compositional}. Video generation models were previously used in reinforcement learning as rewards~\citep{escontrela2024video}, pretraining methods~\citep{tomar2024videooccupancymodels}, for simulating and planning manipulation actions~\citep{finn2017deep,liang2024dreamitate} and for generating paths in indoor environments~\citep{hirose2019deep,koh2021pathdreamer}. Interestingly, diffusion models~\citep{sohl2015deep,ho2020denoising} are useful both for video tasks like generation~\citep{voleti2022mcvd} and prediction~\citep{lin2024veditlatentpredictionarchitecture}, but also for view synthesis~\citep{Chan_2023_ICCV,pooledreamfusion,cho}. Differently, we use a conditional diffusion transformer to simulate trajectories for planning without explicit $3$D representations or priors.

\section{Navigation World Models}

\subsection{Formulation}

Next, we turn to describe our NWM formulation. Intuitively, a NWM is a model that receives the current state of the world (e.g. an image observation) and a navigation action describing where to move and how to rotate. The model then produces the next state of the world with respect to the agent's point of view.

We are given an egocentric video dataset together with agent navigation actions $D = \{(x_0, a_0, ..., x_T, a_T)\}^{n}_{i=1}$, such that $x_i\in \mathbb{R}^{H\times W \times 3}$ is an image and $a_i=(u,\phi)$ is a navigation command given by translation parameter $u\in\mathbb{R}^{2}$ that controls the change in forward/backward and right/left motion, as well as $\phi \in \mathbb{R}$ that controls the change in yaw rotation angle.\footnote{This can be naturally extended to three dimensions by having $u\in\mathbb{R}^{3}$ and $\theta\in\mathbb{R}^3$ defining yaw, pitch and roll. For simplicity, we assume navigation on a flat surface with fixed pitch and roll.}

The navigation actions ${a_i}$ can be fully observed (as in Habitat~\citep{savva2019habitat}), e.g. moving forward towards a wall will trigger a response from the environment based on physics, which will lead to the agent staying in place, whereas in other environments the navigation actions can be approximated based on the change in the agent's location.

Our goal is to learn a world model $F$, a stochastic mapping from previous latent observation(s) $\mathbf{s}_\tau$ and action $a_\tau$ to future latent state representation $s_{t+1}$:
\begin{align}
    \label{eq:basic}
     s_i = \text{enc}_{\theta}(x_{i}) && s_{\tau+1} \sim F_{\theta}(s_{\tau+1}\mid\mathbf{s_\tau}, a_\tau) 
\end{align}
Where $\mathbf{s_\tau}=({s_\tau,..., s_{\tau-m}})$ are the past $m$ visual observations encoded via a pretrained VAE~\citep{blattmann2023stable}. Using a VAE has the benefit of working with compressed latents, allowing to decode predictions back to pixel space for visualization. 

Due to the simplicity of this formulation, it can be naturally shared across environments and easily extended to more complex action spaces, like controlling a robotic arm. Different than~\cite{hafnerdream}, we aim to train a single world model across environments and embodiments, without using task or action embeddings like in~\cite{hansentd}. 

The formulation in Equation~\ref{eq:basic} models action but does not allow control over the temporal dynamics. We extend this formulation with a time shift input $k\in[{T_\text{min}}, {T_\text{max}}]$, setting $a_\tau = (u,\phi, k)$, 
thus now $a_\tau$ specifies the time change $k$, used to determine how many steps should the model move into the future (or past). Hence, given a current state $s_\tau$, we can randomly choose a timeshift $k$ and use the corresponding time shifted video frame as our next state $s_{\tau+1}$. The navigation actions can then be approximated to be a summation from time $\tau$ to $m=\tau+k-1$:
\begin{align}
\label{eq:compose-actions}
    u_{\tau \rightarrow m} = \sum_{t=\tau}^{m} u_t && \phi_{\tau \rightarrow m} = \sum_{t=\tau}^{m}\phi_{t}\mod2\pi
\end{align}
This formulation allows learning both navigation actions, but also the environment temporal dynamics. In practice, we allow time shifts of up to $\pm16$ seconds.

One challenge that may arise is the entanglement of actions and time. For example, if reaching a specific location always occurs at a particular time, the model may learn to rely solely on time and ignore the subsequent actions, or vice versa. In practice, the data may contain natural counterfactuals—such as reaching the same area at different times. To encourage these natural counterfactuals, we sample multiple goals for each state during training. We further explore this approach in Section~\ref{sec:experiments}.

\subsection{Diffusion Transformer as World Model}
\label{sec:diff-transformer}
As mentioned in the previous section, we design $F_{\theta}$ as a stochastic mapping so it can simulate stochastic environments. This is achieved using a Conditional Diffusion Transformer (CDiT) model, described next. 

\begin{figure}
  \centering
\includegraphics[width=0.75\linewidth]{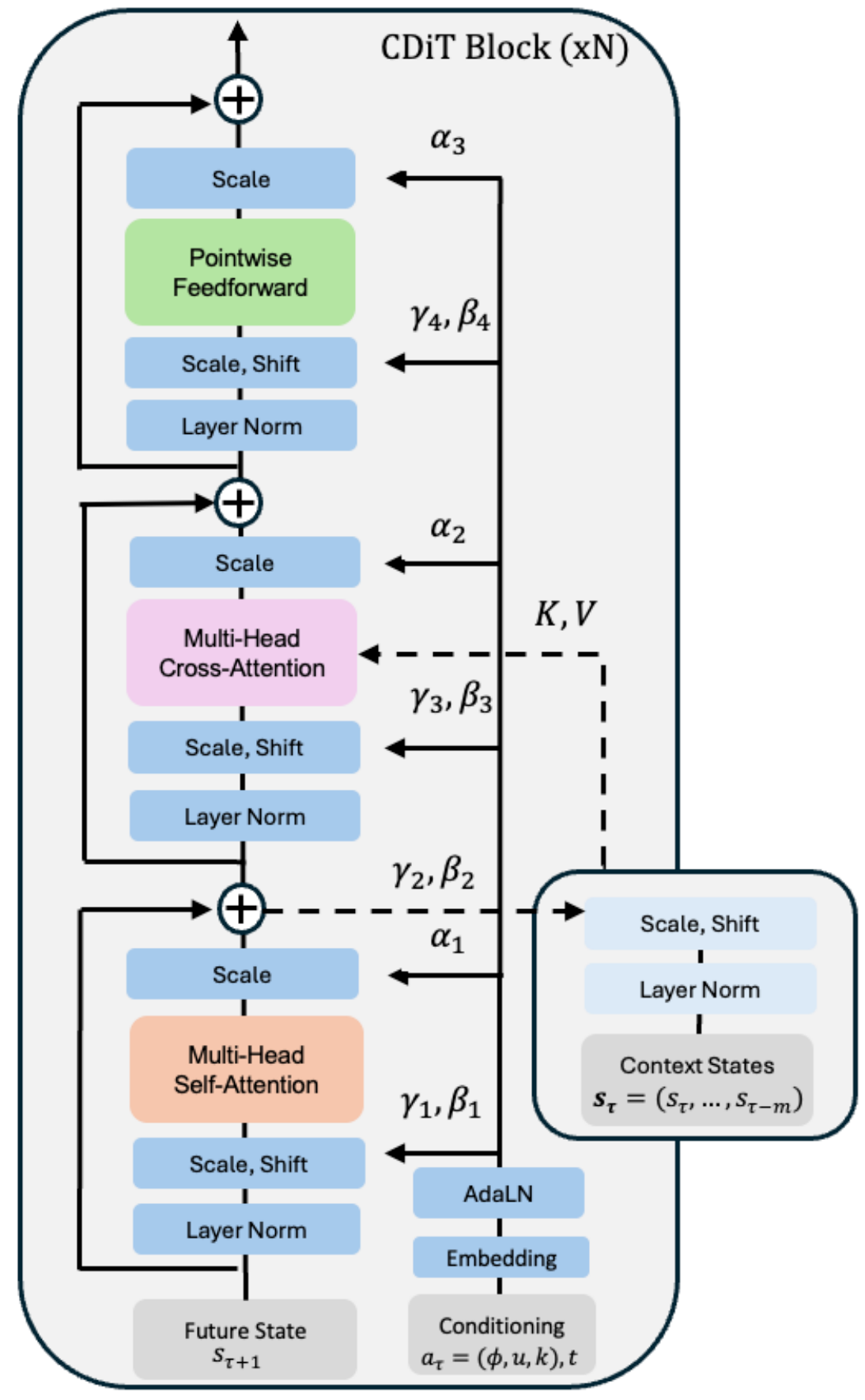}
  \caption{
  \textbf{Conditional Diffusion Transformer (CDiT) Block}. The block’s complexity is linear with the number of frames.}
  \label{fig:arch}
  \vspace{-5mm}
\end{figure}

\vspace{1.3mm}
\noindent\textbf{Conditional Diffusion Transformer Architecture}. The architecture we use is a temporally autoregressive transformer model utilizing the efficient CDiT block (see Figure~\ref{fig:arch}), which is applied $\times N$ times over the input sequence of latents with input action conditioning.

CDiT enables time-efficient autoregressive modeling by constraining the attention in the first attention block only to tokens from the target frame which is being denoised. To condition on tokens from past frames, we incorporate a cross-attention layer, such that every query token from the current target attends to tokens from past frames, which are used as keys and values. The cross-attention then contextualizes the representations using a skip connection layer. 

To condition on the navigation action $a\in\mathbb{R}^3$, we first map each scalar to $\mathbb{R}^\frac{d}{3}$ by extracting sine-cosine features, then applying a $2$-layer $\text{MLP}$, and concatenating them into a single vector  $\psi_a \in \mathbb{R}^d$. We follow a similar process to map the timeshift $k\in\mathbb{R}$ to $\psi_k\in\mathbb{R}^d$  and the diffusion timestep $t\in\mathbb{R}$ to $\psi_k\in\mathbb{R}^d$. Finally we sum all embeddings into a single vector used for conditioning:
\begin{align}
\label{eq:embedding}
\xi = \psi_a + \psi_k + \psi_t
\end{align}
$\xi$ is then fed to an AdaLN~\citep{xu2019understandingimprovinglayernormalization} block to generate scale and shift coefficients that modulate the Layer Normalization~\citep{lei2016layer} outputs, as well as the outputs of the attention layers. To train on unlabeled data, we simply omit explicit navigation actions when computing $\xi$ (see Eq.~\ref{eq:embedding}).

An alternative approach is to simply use DiT~\citep{Peebles_2023_ICCV}, however, applying a DiT on the full input is computationally expensive. Denote $n$ the number of input tokens per frame, and $m$ the number of frames, and $d$ the token dimension. Scaled Multi-head Attention Layer~\citep{vaswani2017attention} complexity is dominated by 
 the attention term $O(m^2n^2d)$, which is quadratic with context length. In contrast, our CDiT block is dominated by the cross-attention layer complexity $O(mn^2d)$, which is linear with respect to the context, allowing us to use longer context size. We analyze these two design choices in Section~\ref{sec:experiments}. CDiT resembles the original Transformer Block~\citep{vaswani2017attention}, without applying expensive self-attention over the context tokens.

\vspace{1.3mm}
\noindent\textbf{Diffusion Training}. In the forward process, noise is added to the target state $s_{\tau+1}$ according to a randomly chosen timestep $t \in \{1, \dots, T\}$. The noisy state $s_{\tau+1}^{(t)}$ can be defined as: $s_{\tau+1}^{(t)} = \sqrt{\alpha_t} s_{\tau+1} + \sqrt{1 - \alpha_t} \epsilon$, where \( \epsilon \sim \mathcal{N}(0, I) \) is Gaussian noise, and \( \{\alpha_t\} \) is a noise schedule controlling the variance. As \( t \) increases, \( s_{\tau+1}^{(t)} \) converges to pure noise. The reverse process attempts to recover the original state representation $s_{\tau+1}$ from the noisy version $s_{\tau+1}^{(t)}$, conditioned on the context $\mathbf{s}_{\tau}$, the current action $a_\tau$, and the diffusion timestep $t$. We define $F_\theta(s_{\tau+1} | \mathbf{s}_{\tau}, a_\tau, t)$ as the denoising neural network model parameterized by $\theta$. We follow the same noise schedule and hyperparams of DiT~\citep{Peebles_2023_ICCV}.

\vspace{1.3mm}
\noindent\textbf{Training Objective}. The model is trained to minimize the mean-squared between the clean and predicted target, aiming to learn the denoising process:
\begin{align*}
\mathcal{L}_\text{simple} = \mathbb{E}_{s_{\tau+1}, a_\tau, \mathbf{s}_{\tau}, \epsilon, t} \left[ \| s_{\tau+1} - F_\theta(s_{\tau+1}^{(t)} | \mathbf{s}_{\tau}, a_\tau, t) \|_2^2 \right].
\end{align*}

In this objective, the timestep $t$ is sampled randomly to ensure that the model learns to denoise frames across varying levels of corruption. By minimizing this loss, the model learns to reconstruct $s_{\tau+1}$ from its noisy version $s_{\tau+1}^{(t)}$, conditioned on the context $\mathbf{s}_{\tau}$ and action $a_{\tau}$, thereby enabling the generation of realistic future frames. Following~\citep{Peebles_2023_ICCV}, we also predict the covariance matrix of the noise and supervise it with the variational lower bound loss $\mathcal{L}_\text{vlb}$~\cite{pmlr-v139-nichol21a}.

\subsection{Navigation Planning with World Models}
\label{sec:methods-navi}
Here we move to describe how to use a trained NWM to plan navigation trajectories. Intuitively, if our world model is familiar with an environment, we can use it to simulate navigation trajectories, and choose the ones which reach the goal. In an unknown, out of distribution environments, long term planning might rely on imagination.

Formally, given the latent encoding $s_0$ and navigation target $s^*$, we look for a sequence of actions $(a_0, ..., a_{T-1})$ that maximizes the likelihood of reaching $s^*$.
Let $\mathcal{S}({s}_T, s^*)$ represent the unnormalized score for reaching state $s^*$ with $s_T$ given the initial condition $s_0$, actions $\mathbf{a}=( a_0, \dots, a_{T-1} )$, and states $\mathbf{s}=({s}_1,\dots {s}_T)$ obtained by autoregressively rolling out the NWM: $\mathbf{s}\sim F_{\theta}(\cdot|s_0,\mathbf{a})$.

We define the energy function $\mathcal{E}(s_0, a_0, \dots, a_{T-1}, s_T)$, such that minimizing the energy corresponds to maximizing the unnormalized perceptual similarity score and following potential constraints on the states and actions:
\begin{align}
\label{eq:score}
\mathcal{E}(s_0, a_0, \dots, a_{T-1}, s_T) = -\mathcal{S}(s_T, s^*) + && \\
\nonumber + \sum_{\tau=0}^{T-1} \mathbb{I}(a_\tau \notin \mathcal{A}_{\text{valid}}) + \sum_{\tau=0}^{T-1} \mathbb{I}(s_\tau \notin \mathcal{S}_{\text{safe}}),
\end{align}
The similarity is computed by decoding $s^*$ and $s_T$ to pixels using a pretrained VAE decoder~\citep{blattmann2023stable} and then measuring the perceptual similarity~\citep{zhang2018perceptual,fu2024dreamsim}. Constraints like ``never go left then right'' can be encoded by constraining $a_\tau$ to be in a valid action set $\mathcal{A}_{\text{valid}}$, and ``never explore the edge of the cliff'' by ensuring such states $s_\tau$ are in $\mathcal{S}_{\text{safe}}$. $\mathbb{I}(\cdot)$ denotes the indicator function that applies a large penalty if any action or state constraint is violated.

The problem then reduces to finding the actions that minimize this energy function:
\begin{equation}
\begin{aligned}
\label{eq:planning-loss}
\arg\min_{a_0, \dots, a_{T-1}} \mathbb{E}_{\mathbf{s}} \left[ \mathcal{E}(s_0, a_0, \dots, a_{T-1}, s_T) \right]  
\end{aligned}
\end{equation}
This objective can be reformulated as a Model Predictive Control (MPC) problem, and we optimize it using the Cross-Entropy Method~\citep{rubinstein1997optimization}, a simple derivative-free and population-based optimization method which was recently used with with world models for planning~\citep{zhou2024dinowmworldmodelspretrained}. We include an overview of the Cross-Entropy Method and the full optimization technical details in Appendix~\ref{sec:supp:optimization}.

\vspace{1.3mm}
\noindent\textbf{Ranking Navigation Trajectories}. Assuming we have an existing navigation policy~$\Pi(\mathbf{a}|s_0, s^*)$, we can use NWMs to rank sampled trajectories. Here we use NoMaD~\citep{sridhar2024nomad}, a state-of-the-art navigation policy for robotic navigation. To rank trajectories, we draw multiple samples from $\Pi$ and choose the one with the lowest energy, like in Eq.~\ref{eq:planning-loss}.
\section{Experiments and Results}
\label{sec:experiments}
We describe the experimental setting, our design choices, and compare NWM to previous approaches. Additional results are included in the Supplementary Material.

\begin{figure*}[t]
    \centering
        \href{https://www.amirbar.net/nwm/index.html#baselines-ablation}{\includegraphics[width=1\linewidth]{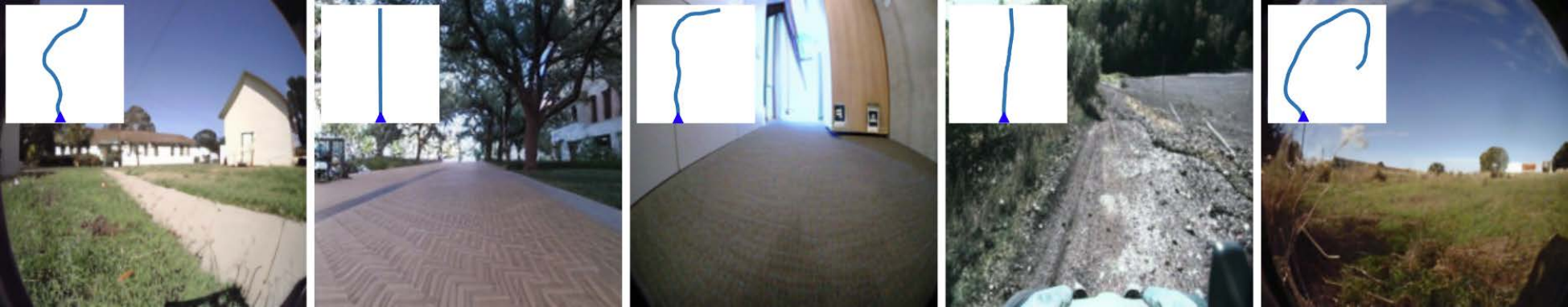}}
    \caption{\textbf{Following trajectories in known environments.} We include qualitative video generation comparisons of different models following ground truth trajectories. \textbf{Click on the image to play the video clip in a browser}.}
    \label{fig:ablations}
\end{figure*}
\begin{figure*}[t]
    \centering
    \begin{minipage}{0.48\textwidth}
        \small
        \centering
        \resizebox{\linewidth}{!}{%
        \begin{tabular}{l|ccccc}
        \hline
        ablation & lpips $\downarrow$ & dreamsim $\downarrow$ & psnr $\uparrow$  \\
        \hline
        1 & $0.312 \pm 0.001$ & $0.098 \pm 0.001$ & $15.044 \pm 0.031$ \\ 
        2~~~\#goals & $0.305 \pm 0.000$ & $0.096 \pm 0.001$ & $15.154 \pm 0.017$ \\ 
        4 & $\textbf{0.296}$ $\pm 0.002$ & $\textbf{0.091}$ $\pm 0.001$ & $\textbf{15.331}$ $\pm 0.027$ \\ 
        \hline
        1 & $0.304 \pm 0.001$ & $0.097 \pm 0.001$ & $15.223 \pm 0.033$ \\ 
        2~~~\#context& $0.302 \pm 0.001$ & $0.095 \pm 0.000$ & $15.274 \pm 0.027$ \\ 
        4 & $\textbf{0.296}$ $\pm 0.002$ & $\textbf{0.091}$ $\pm 0.001$ & $\textbf{15.331}$ $\pm 0.027$ \\ 
        \hline
        time only & $0.760 \pm 0.001$ & $0.783 \pm 0.000$ & $7.839 \pm 0.017$ \\
        action only & $0.318 \pm 0.002$ & $0.100 \pm 0.000$ & $14.858 \pm 0.055$ \\
        action + time & $\textbf{0.295}$ $\pm 0.002$ & $\textbf{0.091}$ $\pm 0.001$ & $\textbf{15.343}$ $\pm 0.060$ \\
        \hline
        
        \end{tabular}
        }%
        \captionof{table}{\textbf{Ablations} of predicted goals per sample number, context size, and the use of action and time conditioning. We report prediction results $4$ seconds into the future on RECON.}
        \label{tab:ablations}
        \vspace{0.5cm} 
    \centering
    \includegraphics[width=1\linewidth]{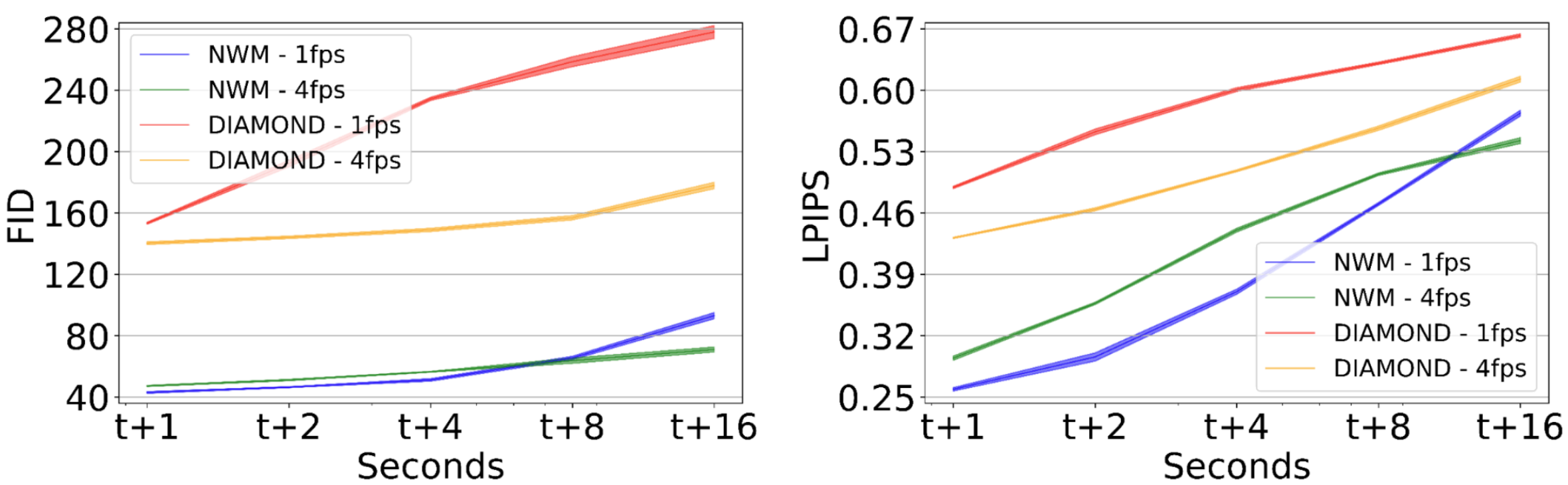}
    \captionof{figure}{\textbf{Comparing generation accuracy and quality} of NWM and DIAMOND at $1$ and $4$ FPS as function of time, up to $16$ seconds of generated video on the RECON dataset.}
    \label{fig:traj-follow}
    \end{minipage}%
            \hspace{0.02\textwidth} 
    \begin{minipage}{0.49\textwidth}
            \centering
            \vspace{-2mm}
            \includegraphics[width=0.95\linewidth]{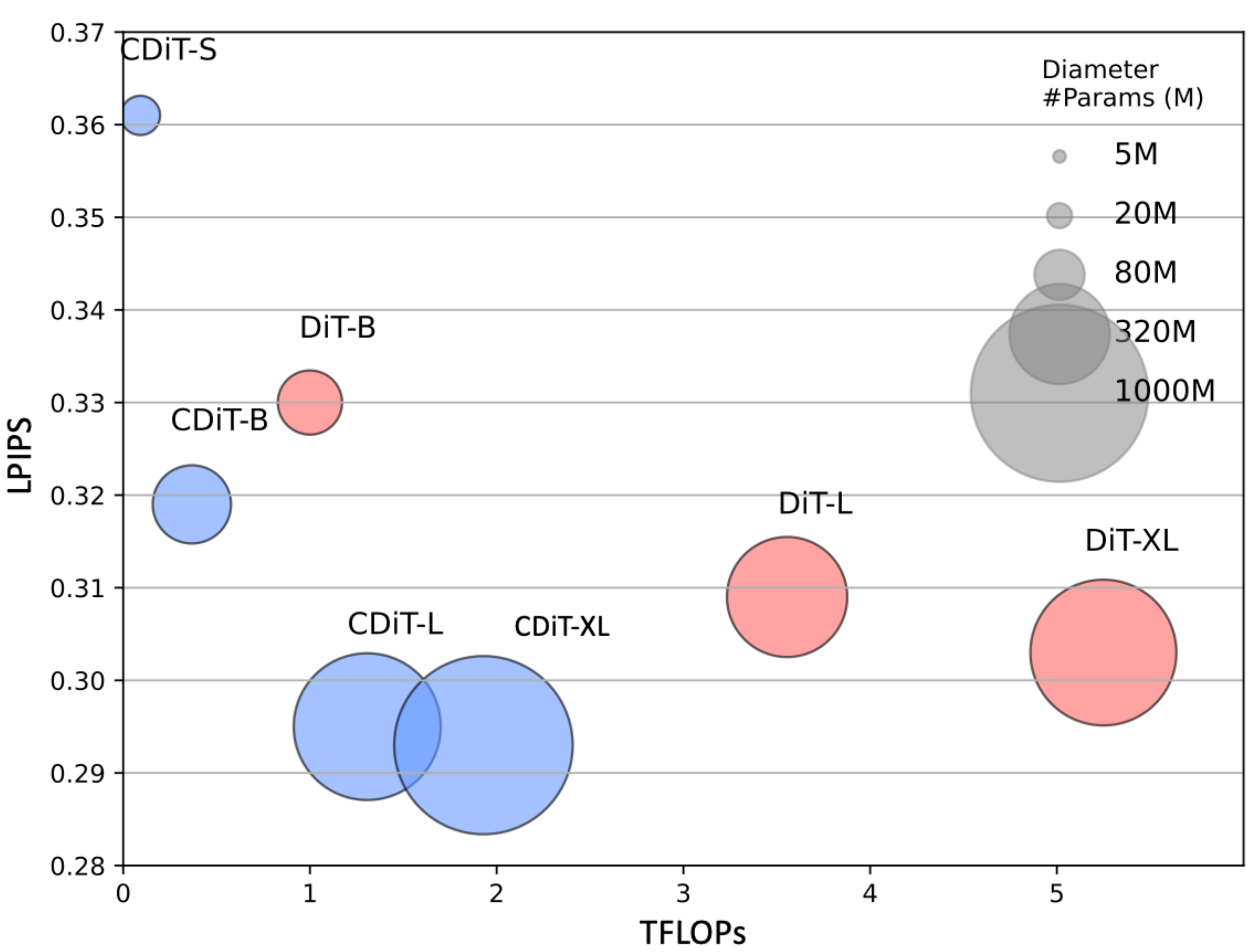}
            \vspace{-1mm}
            \caption{\textbf{CDiT vs. DiT}. Measuring how well models predict $4$ seconds into the future on RECON. We report LPIPS as a function of Tera FLOPs, lower is better.}
            \label{fig:scale}
    \label{fig:baseline}
\vspace{0.03\textwidth}
\small
\centering
\begin{tabular}{l|ccccc}
\hline
model & diamond & NWM (ours) \\
\hline
 FVD $\downarrow$ & $762.734 \pm 3.361$ & $\textbf{200.969}$ $\pm 5.629$\\
\hline
\end{tabular}
\caption{\textbf{Comparison of Video Synthesis Quality.} $16$ second videos generated at 4 FPS on RECON.}
\label{tab:fvd}
    \end{minipage}


\end{figure*}

\subsection{Experimental Setting}
\noindent\textbf{Datasets.} For all robotics datasets (SCAND~\citep{karnan2022socially}, TartanDrive~\citep{triest2022tartandrive}, RECON~\citep{shah2021rapid}, and HuRoN~\citep{hirose2023sacson}), we have access to the location and rotation of robots, allowing us to infer relative actions compare to current location (see Eq.~\ref{eq:compose-actions}). To standardize the step size across agents, we divide the distance agents travel between frames by their average step size in meters, ensuring the action space is similar for different agents. We further filter out backward movements, following NoMaD~\citep{sridhar2024nomad}. Additionally, we use unlabeled Ego4D~\citep{grauman2022ego4d} videos, where the only action we consider is 
time shift. SCAND provides video footage of socially compliant navigation in diverse environments, TartanDrive focuses on off-road driving, RECON covers open-world navigation, HuRoN captures social interactions. We train on unlabeled Ego4D videos and GO Stanford~\citep{hirose2018gonet} serves as an unknown evaluation environment. For the full details, see Appendix~\ref{supp:sec:exp-study}.

\vspace{1.3mm}
\noindent\textbf{Evaluation Metrics.} We evaluate predicted navigation trajectories using Absolute Trajectory Error (ATE) for accuracy and Relative Pose Error (RPE) for pose consistency~\citep{sturm2012evaluating}. To check how semantically similar are world model predictions to ground truth images, we apply LPIPS~\citep{zhang2018unreasonable} and DreamSim~\citep{fu2024dreamsim}, measuring perceptual similarity by comparing deep features, and PSNR for pixel-level quality. For image and video synthesis quality, we use FID~\citep{heusel2017gans} and FVD~\citep{unterthiner2019fvd} which evaluate the generated data distribution. See Appendix~\ref{supp:sec:exp-study} for more details.

\vspace{1.3mm}
\noindent\textbf{Baselines.} We consider all the following baselines.
\vspace{1.3mm}
\begin{itemize}
    \item \textbf{DIAMOND}~\citep{alonso2024diffusionworldmodelingvisual} is a diffusion world model based on the UNet~\citep{ronneberger2015u} architecture. We use DIAMOND in the offline-reinforcement learning setting following their public code. The diffusion model is trained to autoregressively predict at $56$x$56$ resolution alongside an upsampler to obtrain $224$x$224$ resolution predictions. To condition on continuous actions, we use a linear embedding layer. 
    \item \textbf{GNM}~\citep{shah2023gnm} is a general goal-conditioned navigation policy trained on a dataset soup of robotic navigation datasets with a fully connected trajectory prediction network. GNM is trained on multiple datasets including SCAND, TartanDrive, GO Stanford, and RECON. 
    \item \textbf{NoMaD}~\citep{sridhar2024nomad} extends GNM using a diffusion policy for predicting trajectories for robot exploration and visual navigation. NoMaD is trained on the same datasets used by GNM and on HuRoN.
\end{itemize}

\vspace{1.3mm}
\noindent\textbf{Implementation Details.} In the default experimental setting we use a CDiT-XL of $1B$ parameters with context of $4$ frames, a total batch size of $1024$, and $4$ different navigation goals, leading to a final total batch size of $4096$. We use the Stable Diffusion~\citep{blattmann2023stable} VAE tokenizer, similar as in DiT~\citep{Peebles_2023_ICCV}. We use the AdamW~\citep{loshchilov2017decoupled} optimizer with a learning rate of $8e-5$. After training, we sample $5$ times from each model to report mean and std results. XL sized model are trained on $8$ H100 machines, each with $8$ GPUs.
Unless otherwise mentioned, we use the same setting as in DiT-*/2 models.

\subsection{Ablations}
Models are evaluated on single-step $4$ seconds future prediction on validation set trajectories on the known environment RECON. We evaluate the performance against the ground truth frame by measuring LPIPS, DreamSim, and PSNR. We provide qualitative examples in Figure~\ref{fig:ablations}.

\vspace{1.3mm}
\noindent\textbf{Model Size and CDiT}. We compare CDiT (see Section~\ref{sec:diff-transformer}) with a standard DiT in which all context tokens are fed as inputs. We hypothesize that for navigating known environments, the capacity of the model is the most important, and the results in Figure~\ref{fig:scale}, indicate that CDiT indeed performs better with models of up to $1$B parameters, while consuming less than $2\times$ FLOPs. Surprisingly, even with equal amount of parameters (e.g, CDiT-L compared to DiT-XL), CDiT is $4\times$ faster and performs better. 

\vspace{1.3mm}
\noindent\textbf{Number of Goals}. We train models with variable number of goal states given a fixed context, changing the number of goals from $1$ to $4$. Each goal is randomly chosen between $\pm16$ seconds window around the current state. The results reported in Table~\ref{tab:ablations} indicate that using $4$ goals leads to significantly improved prediction performance in all metrics.

\vspace{1.3mm}
\noindent\textbf{Context Size}. We train models while varying the number of conditioning frames from $1$ to $4$ (see Table~\ref{tab:ablations}). Unsurprisingly, more context helps, and with short context the model often ``lose track'', leading to poor predictions.

\vspace{1.3mm}
\noindent\textbf{Time and Action Conditioning}. We train our model with both time and action conditioning and test how much each input contributes to the prediction performance (we include the results in Table~\ref{tab:ablations}. We find that running the model with time only leads to poor performance, while not conditioning on time leads to small drop in performance as well. This confirms that both inputs are beneficial to the model.

\begin{figure*}[t]
    \centering
    \href{https://www.amirbar.net/nwm/index.html#ranking}{
    {\includegraphics[width=1\linewidth]{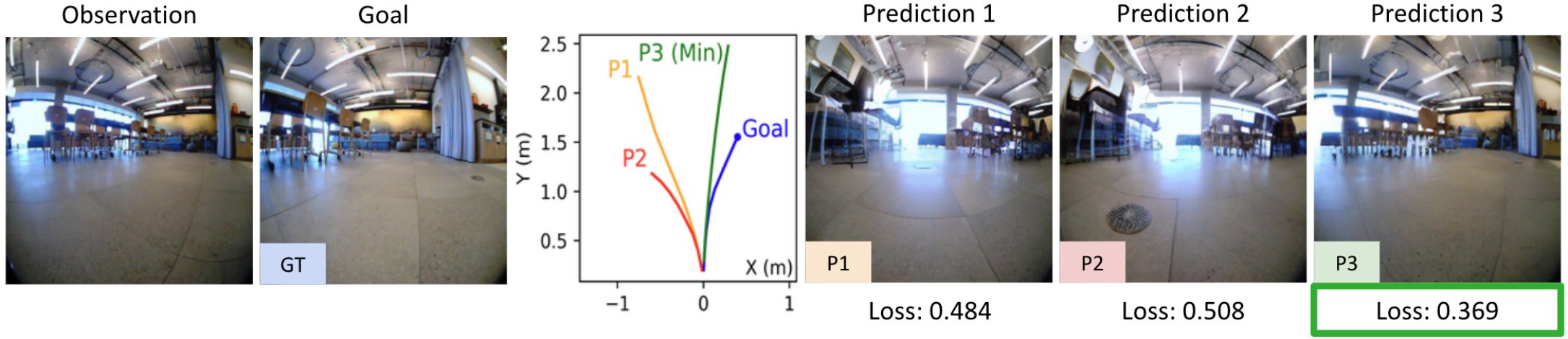}}}
    \vspace{-7mm}
    \caption{\textbf{Ranking an external policy's trajectories using NWM.} To navigate from the observation image to the goal, we sample trajectories from NoMaD~\citep{sridhar2024nomad}, simulate each of these trajectories using NWM, score them (see Equation~\ref{eq:score}), and rank them. With NWM we can accurately choose trajectories that are closer to the groundtruth trajectory. \textbf{Click the image to play examples in a browser}.}
    \label{fig:nomad}
\end{figure*}
\begin{table*}[htbp]
    \begin{minipage}{0.45\textwidth}
        \footnotesize
        \centering
        \begin{tabular}{l|ccc}
        \hline
        model & ATE $\downarrow$ & RPE $\downarrow$ \\
        \hline
        GNM & $1.87$ $\pm$ $0.00$ & $0.73$ $\pm$ $0.00$ \\
        NoMaD &  $1.93$ $\pm$ $0.04$ & $0.52$ $\pm$ $0.00$ \\
        \hline
        NWM + NoMaD ($\times 16$) & $1.83$ $\pm$ $0.03$ & $0.50$ $\pm$ $0.01$ \\ 
        NWM + NoMaD ($\times 32$) & $1.78$ $\pm$ $0.03$ & $0.48$ $\pm$ $0.01$ \\  
        \hline
        NWM (planning) & $\textbf{1.13}$ $\pm$ $0.02$ & $\textbf{0.35}$ $\pm$ $0.01$ \\  
        \hline
        \end{tabular}
        \caption{\textbf{Goal Conditioned Visual Navigation}. ATE and RPE results on RECON, predicting $2$ second trajectories. NWM achieves improved results on all metrics compared to previous approaches NoMaD~\citep{sridhar2024nomad} and GNM~\citep{shah2023gnm}.}
        \label{tab:traj-follow}
            \end{minipage} %
    \hspace{0.1\textwidth} 
    \begin{minipage}{0.45\textwidth}
        \small
        \centering
        \begin{tabular}{l|cc}
        \hline
        model & Rel. $\delta u$  $\downarrow$ & Rel. $\delta \phi$ $\downarrow$ \\
        \hline
        forward first & $+0.36 \pm 0.01$ & $+0.61 \pm 0.02$ \\
        left-right first & $-0.03 \pm 0.01$ & $+0.20 \pm 0.01$ \\
        straight then forward & $+0.08 \pm 0.01$ & $+0.22 \pm 0.01$ \\
        \hline
        \end{tabular}
        \caption{
        \textbf{Planning with Navigation Constraints.} We present results for planning with NWM under three action constraints, reporting the differences in final position ($\delta u$) and yaw ($\delta \phi$) relative to the no-constraints baseline. All constraints are met, demonstrating that NWM can effectively adhere to them.}
        \label{tab:plan_constraints}
        \end{minipage}
\end{table*}

\subsection{Video Prediction and Synthesis}
We evaluate how well our model follows ground truth actions and predicts future states. The model is conditioned on the first image and context frames, then autoregressively predicts the next state using ground truth actions, feeding back each prediction. We compare predictions to ground truth images at $1$, $2$, $4$, $8$, and $16$ seconds, reporting FID and LPIPS on the RECON dataset. Figure~\ref{fig:traj-follow} shows performance over time compared to DIAMOND at $4$ FPS and $1$ FPS, showing that NWM predictions are significantly more accurate than DIAMOND. Initially, the NWM $1$ FPS variant performs better, but after $8$ seconds, predictions degrade due to accumulated errors and loss of context and the $4$ FPS becomes superior. See qualitative examples in Figure~\ref{fig:ablations}.

\vspace{1.3mm}
\noindent\textbf{Generation Quality.} To evaluate video quality, we auto-regressively predict videos at $4$ FPS for $16$ seconds to create videos, while conditioning on ground truth actions. We then evaluate the quality of videos generated using FVD, compared to DIAMOND~\citep{alonso2024diffusionworldmodelingvisual}. The results in Figure~\ref{tab:fvd} indicate that NWM outputs higher quality videos.

\begin{figure*}[ht!]
    \centering
\href{https://www.amirbar.net/nwm/index.html#unknown-environments}{    \includegraphics[width=1\linewidth]{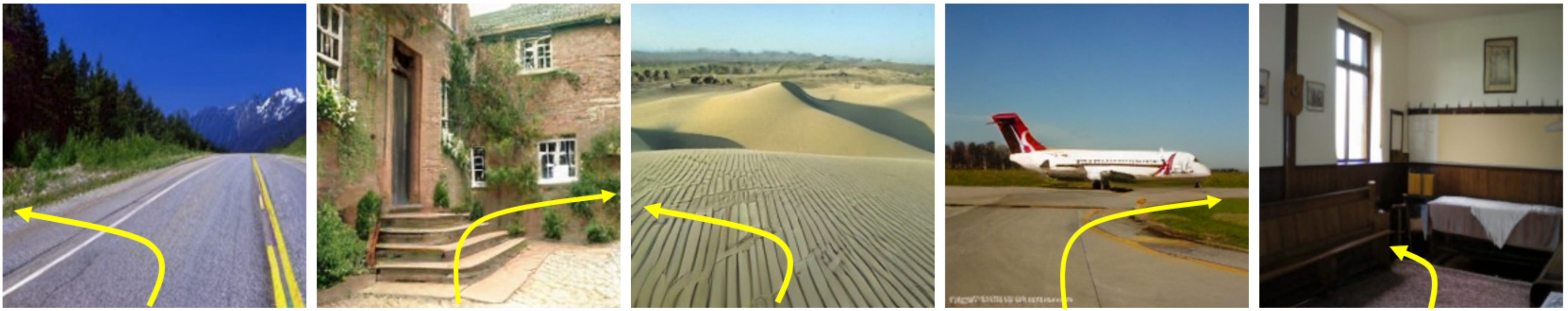}}
    \caption{\textbf{Navigating Unknown Environments}. NWM is conditioned on a single image, and autoregressively predicts the next states given the associated actions (marked in yellow). \textbf{Click on the image to play the video clip in a browser}. }
    \label{fig:out_of_domain}
\end{figure*}
\begin{table*}[ht!]
\footnotesize
\centering
\begin{tabular}{l|cccccccccc}
\hline
data & \multicolumn{3}{c}{unknown environment (Go Stanford)} & \multicolumn{3}{c}{known environment (RECON)}\\
 & lpips $\downarrow$ & dreamsim $\downarrow$ & psnr $\uparrow$ & lpips $\downarrow$ & dreamsim $\downarrow$ & psnr $\uparrow$ \\
\hline
in-domain data & $0.658 \pm 0.002$ & $0.478 \pm 0.001$ & $11.031 \pm 0.036$ &  $\textbf{0.295}$ $\pm 0.002$ & $\textbf{0.091}$ $\pm 0.001$ & $\textbf{15.343}$ $\pm 0.060$\\
+ Ego4D~\text{\footnotesize(unlabeled)} & $\textbf{0.652}$ $\pm 0.003$ & $\textbf{0.464}$ $\pm 0.003$ & $\textbf{11.083}$ $\pm 0.064$ & $0.368 \pm 0.003$ & $0.138 \pm 0.002$ & $14.072 \pm 0.075$ \\ 
\hline
\end{tabular}
\caption{\textbf{Training on additional unlabeled data improves performance on  unseen environments. } Reporting results on unknown environment (Go Stanford) and known one (RECON). Results reported by evaluating $4$ seconds into the future.}
\label{tab:oodm}
\end{table*}

\subsection{Planning Using a Navigation World Model}
Next, we turn to describe experiments that measure how well can we navigate using a NWM. We include the full technical details of the experiments in Appendix~\ref{sec:supp:results}.

\vspace{1.3mm}
\noindent\textbf{Standalone Planning.} We demonstrate that NWM can be effectively used independently for goal-conditioned navigation. We condition it on past observations and a goal image, and use the Cross-Entropy Method to find a trajectory that minimizes the LPIPS similarity of the last predicted image to the goal image (see Equation~\ref{eq:planning-loss}). To rank an action sequence, we execute the NWM and measure LPIPS between the last state and the goal $3$ times to get an average score. We generate trajectories of length $8$, with temporal shift of $k=0.25$. We evaluate the model performance in Table~\ref{tab:traj-follow}. We find that using a NWM for planning leads to competitive results with state-of-the-art policies.

\begin{figure}
    \centering
    \vspace{-6mm}
{\includegraphics[width=1\linewidth]{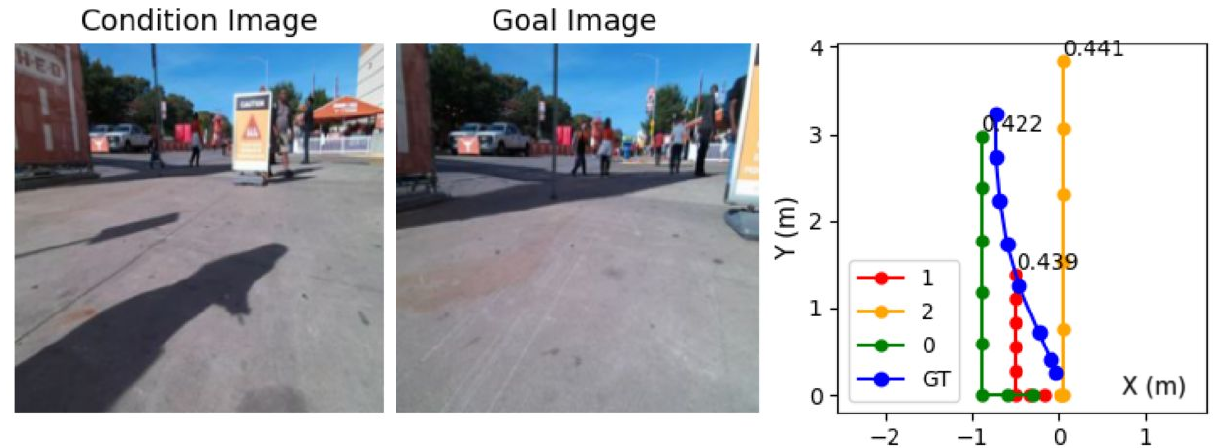}\vspace{-1mm}}
    \caption{
    \textbf{Planning with Constraints Using NWM.} We visualize trajectories planned with NWM under the constraint of moving left or right first, followed by forward motion. The planning objective is to reach the same final position and orientation as the ground truth (GT) trajectory. Shown are the costs for proposed trajectories $0$, $1$, and $2$, with trajectory $0$ (in green) achieving the lowest cost.}
    \label{fig:plan_constraints}
    \vspace{-5mm}
\end{figure}

\vspace{1.3mm}
\noindent\textbf{Planning with Constraints.} World models allow planning under constraints—for example, requiring straight motion or a single turn. We show that NWM supports constraint-aware planning. In \emph{forward-first}, the agent moves forward for 5 steps, then turns for 3. In \emph{left-right first}, it turns for 3 steps before moving forward. In \emph{straight then forward}, it moves straight for 3 steps, then forward. Constraints are enforced by zeroing out specific actions; e.g., in \emph{left-right first}, forward motion is zeroed for the first 3 steps, and Standalone Planning optimizes the rest. We report the norm of the difference in final position and yaw relative to unconstrained planning. Results (Table~\ref{tab:plan_constraints}) show NWM plans effectively under constraints, with only minor performance drops (see examples in Figure~\ref{fig:plan_constraints}).

\vspace{1.3mm}
\noindent\textbf{Using a Navigation World Model for Ranking}. NWM can enhance existing navigation policies in a goal-conditioned navigation. Conditioning NoMaD on past observations and a goal image, we sample $n \in \{16,32\}$ trajectories, each of length $8$, and evaluate them by autoregressively following the actions using NWM. Finally, we rank each trajectory's final prediction by measuring LPIPS similarity with the goal image (see Figure~\ref{fig:nomad}). We report ATE and RPE on all in-domain datasets (Table~\ref{tab:traj-follow}) and find that NWM-based trajectory ranking improves navigation performance, with more samples yielding better results.

\vspace{-1mm}
\subsection{Generalization to Unknown Environments}
Here we experiment with adding unlabeled data, and ask whether NWM can make predictions in new environments using imagination. In this experiment, we train a model on all in-domain datasets, as well as a susbet of unlabeled videos from Ego4D, where we only have access to the time-shift action. We train a CDiT-XL model and test it on the Go Stanford dataset as well as other random images. We report the results in Table~\ref{tab:oodm}, finding that training on unlabeled data leads to significantly better video predictions according to all metrics, including improved generation quality. We include qualitative examples in Figure~\ref{fig:out_of_domain}. Compared to in-domain (Figure \ref{fig:ablations}), the model breaks faster and expectedly hallucinates paths as it generates traversals of imagined environments.
\section{Limitations}
\begin{figure}
    \centering
    \vspace{-5mm}
    \href{https://www.amirbar.net/nwm/index.html#limitations}{
    \includegraphics[width=1\linewidth]{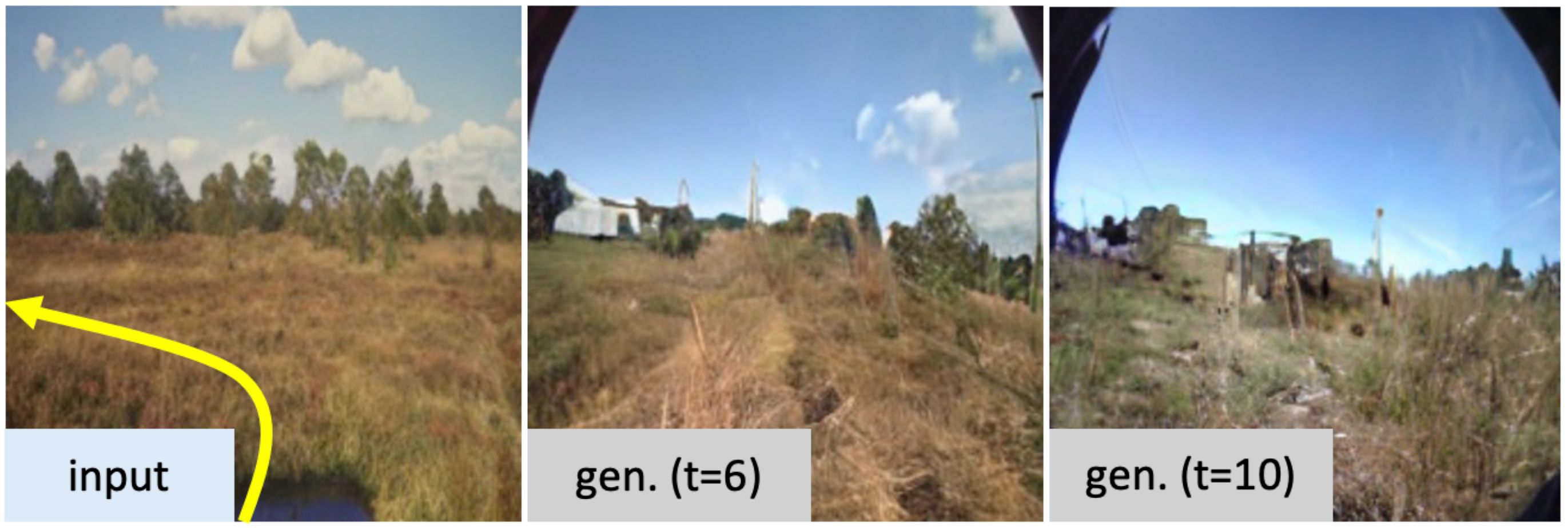}}
    \caption{\textbf{Limitations and Failure Cases.} In unknown environments, a common failure case is mode collapse, where the model outputs slowly become more similar to data seen in training. \textbf{Click on the image to play the video clip in a browser}.
    \vspace{-4mm}
    }
    \label{fig:limitations}
\end{figure}
We identify multiple limitations. First, when applied to out of distribution data, the model tends to slowly lose context and generates next states that resemble the training data, a phenomena that was observed in image generation and is known as mode collapse~\citep{thanh2020catastrophic,srivastava2017veegan}. We include such an example in Figure~\ref{fig:limitations}. Second, while the model can plan, it struggles with simulating temporal dynamics like pedestrian motion (although in some cases it does). Both limitations are likely to be solved with longer context and more training data. Additionally, the model currently utilizes $3$ DoF navigation actions, but extending to $6$ DoF navigation and potentially more (like controlling the joints of a robotic arm) are possible as well, which we leave for future work. 

\section{Discussion}
Our proposed Navigation World Model (NWM) offers a scalable, data-driven approach to learning world models for visual navigation; However, we are not exactly sure yet what representations enable this, as our NWM does not explicitly utilize a structured map of the environment. One idea, is that next frame prediction from an egocentric point of view can drive the emergence of allocentric representations~\cite{Uria2020.11.11.378141}. Ultimately, our approach bridges learning from video, visual navigation, and model-based planning and could potentially open the door to self-supervised systems that not only perceive but can also plan to inform action.

\vspace{3mm}
\noindent\textbf{Acknowledgments.}  We thank Noriaki Hirose for his help with the HuRoN dataset and for sharing his insights, and to Manan Tomar, David Fan, Sonia Joseph, Angjoo Kanazawa, Ethan Weber, Nicolas Ballas, and the anonymous reviewers for their helpful discussions and feedback.

\clearpage
\pagebreak
{
    \small
    \bibliographystyle{ieeenat_fullname}
    \bibliography{main}

\begin{thebibliography}{77}
\providecommand{\natexlab}[1]{#1}
\providecommand{\url}[1]{\texttt{#1}}
\expandafter\ifx\csname urlstyle\endcsname\relax
  \providecommand{\doi}[1]{doi: #1}\else
  \providecommand{\doi}{doi: \begingroup \urlstyle{rm}\Url}\fi

\bibitem[Alonso et~al.()Alonso, Jelley, Micheli, Kanervisto, Storkey, Pearce, and Fleuret]{alonso2024diffusionworldmodelingvisual}
Eloi Alonso, Adam Jelley, Vincent Micheli, Anssi Kanervisto, Amos Storkey, Tim Pearce, and François Fleuret.
\newblock Diffusion for world modeling: Visual details matter in atari.
\newblock In \emph{Thirty-eighth Conference on Neural Information Processing Systems}.

\bibitem[Bar et~al.(2021)Bar, Herzig, Wang, Rohrbach, Chechik, Darrell, and Globerson]{bar2021compositional}
Amir Bar, Roei Herzig, Xiaolong Wang, Anna Rohrbach, Gal Chechik, Trevor Darrell, and Amir Globerson.
\newblock Compositional video synthesis with action graphs.
\newblock In \emph{International Conference on Machine Learning}, pages 662--673. PMLR, 2021.

\bibitem[Bar-Tal et~al.(2024)Bar-Tal, Chefer, Tov, Herrmann, Paiss, Zada, Ephrat, Hur, Liu, Raj, et~al.]{bar2024lumiere}
Omer Bar-Tal, Hila Chefer, Omer Tov, Charles Herrmann, Roni Paiss, Shiran Zada, Ariel Ephrat, Junhwa Hur, Guanghui Liu, Amit Raj, et~al.
\newblock Lumiere: A space-time diffusion model for video generation.
\newblock \emph{arXiv preprint arXiv:2401.12945}, 2024.

\bibitem[Blattmann et~al.(2023)Blattmann, Dockhorn, Kulal, Mendelevitch, Kilian, Lorenz, Levi, English, Voleti, Letts, et~al.]{blattmann2023stable}
Andreas Blattmann, Tim Dockhorn, Sumith Kulal, Daniel Mendelevitch, Maciej Kilian, Dominik Lorenz, Yam Levi, Zion English, Vikram Voleti, Adam Letts, et~al.
\newblock Stable video diffusion: Scaling latent video diffusion models to large datasets.
\newblock \emph{arXiv preprint arXiv:2311.15127}, 2023.

\bibitem[Brooks et~al.(2024)Brooks, Peebles, Holmes, DePue, Guo, Jing, Schnurr, Taylor, Luhman, Luhman, et~al.]{brooks2024video}
Tim Brooks, Bill Peebles, Connor Holmes, Will DePue, Yufei Guo, Li Jing, David Schnurr, Joe Taylor, Troy Luhman, Eric Luhman, et~al.
\newblock Video generation models as world simulators, 2024.

\bibitem[Bruce et~al.(2024)Bruce, Dennis, Edwards, Parker-Holder, Shi, Hughes, Lai, Mavalankar, Steigerwald, Apps, et~al.]{bruce2024genie}
Jake Bruce, Michael~D Dennis, Ashley Edwards, Jack Parker-Holder, Yuge Shi, Edward Hughes, Matthew Lai, Aditi Mavalankar, Richie Steigerwald, Chris Apps, et~al.
\newblock Genie: Generative interactive environments.
\newblock In \emph{Forty-first International Conference on Machine Learning}, 2024.

\bibitem[Chan et~al.(2023)Chan, Nagano, Chan, Bergman, Park, Levy, Aittala, De~Mello, Karras, and Wetzstein]{Chan_2023_ICCV}
Eric~R. Chan, Koki Nagano, Matthew~A. Chan, Alexander~W. Bergman, Jeong~Joon Park, Axel Levy, Miika Aittala, Shalini De~Mello, Tero Karras, and Gordon Wetzstein.
\newblock Generative novel view synthesis with 3d-aware diffusion models.
\newblock In \emph{Proceedings of the IEEE/CVF International Conference on Computer Vision (ICCV)}, pages 4217--4229, 2023.

\bibitem[Chaplot et~al.()Chaplot, Gandhi, Gupta, Gupta, and Salakhutdinov]{chaplotlearning}
Devendra~Singh Chaplot, Dhiraj Gandhi, Saurabh Gupta, Abhinav Gupta, and Ruslan Salakhutdinov.
\newblock Learning to explore using active neural slam.
\newblock In \emph{International Conference on Learning Representations}.

\bibitem[Chen et~al.()Chen, Gupta, and Gupta]{chenlearning}
Tao Chen, Saurabh Gupta, and Abhinav Gupta.
\newblock Learning exploration policies for navigation.
\newblock In \emph{International Conference on Learning Representations}.

\bibitem[Escontrela et~al.(2024)Escontrela, Adeniji, Yan, Jain, Peng, Goldberg, Lee, Hafner, and Abbeel]{escontrela2024video}
Alejandro Escontrela, Ademi Adeniji, Wilson Yan, Ajay Jain, Xue~Bin Peng, Ken Goldberg, Youngwoon Lee, Danijar Hafner, and Pieter Abbeel.
\newblock Video prediction models as rewards for reinforcement learning.
\newblock \emph{Advances in Neural Information Processing Systems}, 36, 2024.

\bibitem[Finn and Levine(2017)]{finn2017deep}
Chelsea Finn and Sergey Levine.
\newblock Deep visual foresight for planning robot motion.
\newblock In \emph{2017 IEEE International Conference on Robotics and Automation (ICRA)}, pages 2786--2793. IEEE, 2017.

\bibitem[Frantar et~al.(2022)Frantar, Ashkboos, Hoefler, and Alistarh]{frantar2022gptq}
Elias Frantar, Saleh Ashkboos, Torsten Hoefler, and Dan Alistarh.
\newblock Gptq: Accurate post-training quantization for generative pre-trained transformers.
\newblock \emph{arXiv preprint arXiv:2210.17323}, 2022.

\bibitem[Frey et~al.(2023)Frey, Mattamala, Chebrolu, Cadena, Fallon, and Hutter]{frey2023fast}
J Frey, M Mattamala, N Chebrolu, C Cadena, M Fallon, and M Hutter.
\newblock Fast traversability estimation for wild visual navigation.
\newblock \emph{Robotics: Science and Systems Proceedings}, 19, 2023.

\bibitem[Fu et~al.(2024)Fu, Tamir, Sundaram, Chai, Zhang, Dekel, and Isola]{fu2024dreamsim}
Stephanie Fu, Netanel Tamir, Shobhita Sundaram, Lucy Chai, Richard Zhang, Tali Dekel, and Phillip Isola.
\newblock Dreamsim: Learning new dimensions of human visual similarity using synthetic data.
\newblock \emph{Advances in Neural Information Processing Systems}, 36, 2024.

\bibitem[Fu et~al.(2022)Fu, Kumar, Agarwal, Qi, Malik, and Pathak]{fu2022coupling}
Zipeng Fu, Ashish Kumar, Ananye Agarwal, Haozhi Qi, Jitendra Malik, and Deepak Pathak.
\newblock Coupling vision and proprioception for navigation of legged robots.
\newblock In \emph{Proceedings of the IEEE/CVF Conference on Computer Vision and Pattern Recognition}, pages 17273--17283, 2022.

\bibitem[Gao et~al.(2024)Gao, Yao, and Xu]{pmlr-v235-gao24p}
Junyu Gao, Xuan Yao, and Changsheng Xu.
\newblock Fast-slow test-time adaptation for online vision-and-language navigation.
\newblock In \emph{Proceedings of the 41st International Conference on Machine Learning}, pages 14902--14919. PMLR, 2024.

\bibitem[Girdhar et~al.(2023)Girdhar, Singh, Brown, Duval, Azadi, Rambhatla, Shah, Yin, Parikh, and Misra]{girdhar2023emu}
Rohit Girdhar, Mannat Singh, Andrew Brown, Quentin Duval, Samaneh Azadi, Sai~Saketh Rambhatla, Akbar Shah, Xi Yin, Devi Parikh, and Ishan Misra.
\newblock Emu video: Factorizing text-to-video generation by explicit image conditioning.
\newblock \emph{arXiv preprint arXiv:2311.10709}, 2023.

\bibitem[Grauman et~al.(2022)Grauman, Westbury, Byrne, Chavis, Furnari, Girdhar, Hamburger, Jiang, Liu, Liu, et~al.]{grauman2022ego4d}
Kristen Grauman, Andrew Westbury, Eugene Byrne, Zachary Chavis, Antonino Furnari, Rohit Girdhar, Jackson Hamburger, Hao Jiang, Miao Liu, Xingyu Liu, et~al.
\newblock Ego4d: Around the world in 3,000 hours of egocentric video.
\newblock In \emph{Proceedings of the IEEE/CVF Conference on Computer Vision and Pattern Recognition}, pages 18995--19012, 2022.

\bibitem[Ha and Schmidhuber(2018)]{ha2018world}
David Ha and J{\"u}rgen Schmidhuber.
\newblock World models.
\newblock \emph{arXiv preprint arXiv:1803.10122}, 2018.

\bibitem[Hafner et~al.({\natexlab{a}})Hafner, Lillicrap, Ba, and Norouzi]{hafnerdream}
Danijar Hafner, Timothy Lillicrap, Jimmy Ba, and Mohammad Norouzi.
\newblock Dream to control: Learning behaviors by latent imagination.
\newblock In \emph{International Conference on Learning Representations}, {\natexlab{a}}.

\bibitem[Hafner et~al.({\natexlab{b}})Hafner, Lillicrap, Norouzi, and Ba]{hafnermastering}
Danijar Hafner, Timothy~P Lillicrap, Mohammad Norouzi, and Jimmy Ba.
\newblock Mastering atari with discrete world models.
\newblock In \emph{International Conference on Learning Representations}, {\natexlab{b}}.

\bibitem[Hansen et~al.()Hansen, Su, and Wang]{hansentd}
Nicklas Hansen, Hao Su, and Xiaolong Wang.
\newblock Td-mpc2: Scalable, robust world models for continuous control.
\newblock In \emph{The Twelfth International Conference on Learning Representations}.

\bibitem[Heusel et~al.(2017)Heusel, Ramsauer, Unterthiner, Nessler, and Hochreiter]{heusel2017gans}
Martin Heusel, Hubert Ramsauer, Thomas Unterthiner, Bernhard Nessler, and Sepp Hochreiter.
\newblock Gans trained by a two time-scale update rule converge to a local nash equilibrium.
\newblock \emph{Advances in neural information processing systems}, 30, 2017.

\bibitem[Hirose et~al.(2018)Hirose, Sadeghian, V{\'a}zquez, Goebel, and Savarese]{hirose2018gonet}
Noriaki Hirose, Amir Sadeghian, Marynel V{\'a}zquez, Patrick Goebel, and Silvio Savarese.
\newblock Gonet: A semi-supervised deep learning approach for traversability estimation.
\newblock In \emph{2018 IEEE/RSJ International Conference on Intelligent Robots and Systems (IROS)}, pages 3044--3051. IEEE, 2018.

\bibitem[Hirose et~al.(2019{\natexlab{a}})Hirose, Sadeghian, Xia, Mart{\'\i}n-Mart{\'\i}n, and Savarese]{hirose2019vunet}
Noriaki Hirose, Amir Sadeghian, Fei Xia, Roberto Mart{\'\i}n-Mart{\'\i}n, and Silvio Savarese.
\newblock Vunet: Dynamic scene view synthesis for traversability estimation using an rgb camera.
\newblock \emph{IEEE Robotics and Automation Letters}, 2019{\natexlab{a}}.

\bibitem[Hirose et~al.(2019{\natexlab{b}})Hirose, Xia, Mart{\'\i}n-Mart{\'\i}n, Sadeghian, and Savarese]{hirose2019deep}
Noriaki Hirose, Fei Xia, Roberto Mart{\'\i}n-Mart{\'\i}n, Amir Sadeghian, and Silvio Savarese.
\newblock Deep visual mpc-policy learning for navigation.
\newblock \emph{IEEE Robotics and Automation Letters}, 4\penalty0 (4):\penalty0 3184--3191, 2019{\natexlab{b}}.

\bibitem[Hirose et~al.(2023)Hirose, Shah, Sridhar, and Levine]{hirose2023sacson}
Noriaki Hirose, Dhruv Shah, Ajay Sridhar, and Sergey Levine.
\newblock Sacson: Scalable autonomous control for social navigation.
\newblock \emph{IEEE Robotics and Automation Letters}, 2023.

\bibitem[Ho et~al.(2020)Ho, Jain, and Abbeel]{ho2020denoising}
Jonathan Ho, Ajay Jain, and Pieter Abbeel.
\newblock Denoising diffusion probabilistic models.
\newblock \emph{Advances in neural information processing systems}, 33:\penalty0 6840--6851, 2020.

\bibitem[Ho et~al.(2022)Ho, Chan, Saharia, Whang, Gao, Gritsenko, Kingma, Poole, Norouzi, Fleet, et~al.]{ho2022imagen}
Jonathan Ho, William Chan, Chitwan Saharia, Jay Whang, Ruiqi Gao, Alexey Gritsenko, Diederik~P Kingma, Ben Poole, Mohammad Norouzi, David~J Fleet, et~al.
\newblock Imagen video: High definition video generation with diffusion models.
\newblock \emph{arXiv preprint arXiv:2210.02303}, 2022.

\bibitem[Karnan et~al.(2022)Karnan, Nair, Xiao, Warnell, Pirk, Toshev, Hart, Biswas, and Stone]{karnan2022socially}
Haresh Karnan, Anirudh Nair, Xuesu Xiao, Garrett Warnell, S{\"o}ren Pirk, Alexander Toshev, Justin Hart, Joydeep Biswas, and Peter Stone.
\newblock Socially compliant navigation dataset (scand): A large-scale dataset of demonstrations for social navigation.
\newblock \emph{IEEE Robotics and Automation Letters}, 7\penalty0 (4):\penalty0 11807--11814, 2022.

\bibitem[Koh et~al.(2021)Koh, Lee, Yang, Baldridge, and Anderson]{koh2021pathdreamer}
Jing~Yu Koh, Honglak Lee, Yinfei Yang, Jason Baldridge, and Peter Anderson.
\newblock Pathdreamer: A world model for indoor navigation.
\newblock In \emph{Proceedings of the IEEE/CVF International Conference on Computer Vision}, pages 14738--14748, 2021.

\bibitem[Kondratyuk et~al.()Kondratyuk, Yu, Gu, Lezama, Huang, Schindler, Hornung, Birodkar, Yan, Chiu, et~al.]{kondratyukvideopoet}
Dan Kondratyuk, Lijun Yu, Xiuye Gu, Jose Lezama, Jonathan Huang, Grant Schindler, Rachel Hornung, Vighnesh Birodkar, Jimmy Yan, Ming-Chang Chiu, et~al.
\newblock Videopoet: A large language model for zero-shot video generation.
\newblock In \emph{Forty-first International Conference on Machine Learning}.

\bibitem[Krizhevsky et~al.(2012)Krizhevsky, Sutskever, and Hinton]{krizhevsky2012imagenet}
Alex Krizhevsky, Ilya Sutskever, and Geoffrey~E Hinton.
\newblock Imagenet classification with deep convolutional neural networks.
\newblock \emph{Advances in neural information processing systems}, 25, 2012.

\bibitem[Lei~Ba et~al.(2016)Lei~Ba, Kiros, and Hinton]{lei2016layer}
Jimmy Lei~Ba, Jamie~Ryan Kiros, and Geoffrey~E Hinton.
\newblock Layer normalization.
\newblock \emph{ArXiv e-prints}, pages arXiv--1607, 2016.

\bibitem[Liang et~al.(2024)Liang, Liu, Ozguroglu, Sudhakar, Dave, Tokmakov, Song, and Vondrick]{liang2024dreamitate}
Junbang Liang, Ruoshi Liu, Ege Ozguroglu, Sruthi Sudhakar, Achal Dave, Pavel Tokmakov, Shuran Song, and Carl Vondrick.
\newblock Dreamitate: Real-world visuomotor policy learning via video generation, 2024.

\bibitem[Lin et~al.(2024{\natexlab{a}})Lin, Nagarajan, Ballas, Assran, Komeili, Bansal, and Sinha]{lin2024veditlatentpredictionarchitecture}
Han Lin, Tushar Nagarajan, Nicolas Ballas, Mido Assran, Mojtaba Komeili, Mohit Bansal, and Koustuv Sinha.
\newblock Vedit: Latent prediction architecture for procedural video representation learning, 2024{\natexlab{a}}.

\bibitem[Lin et~al.(2024{\natexlab{b}})Lin, Du, Watkins, Hafner, Abbeel, Klein, and Dragan]{lin2024learningmodelworldlanguage}
Jessy Lin, Yuqing Du, Olivia Watkins, Danijar Hafner, Pieter Abbeel, Dan Klein, and Anca Dragan.
\newblock Learning to model the world with language, 2024{\natexlab{b}}.

\bibitem[Liu et~al.(2023)Liu, Wu, Van~Hoorick, Tokmakov, Zakharov, and Vondrick]{liu2023zero}
Ruoshi Liu, Rundi Wu, Basile Van~Hoorick, Pavel Tokmakov, Sergey Zakharov, and Carl Vondrick.
\newblock Zero-1-to-3: Zero-shot one image to 3d object.
\newblock In \emph{Proceedings of the IEEE/CVF international conference on computer vision}, pages 9298--9309, 2023.

\bibitem[Loshchilov(2017)]{loshchilov2017decoupled}
I Loshchilov.
\newblock Decoupled weight decay regularization.
\newblock \emph{arXiv preprint arXiv:1711.05101}, 2017.

\bibitem[Mildenhall et~al.(2021)Mildenhall, Srinivasan, Tancik, Barron, Ramamoorthi, and Ng]{mildenhall2021nerf}
Ben Mildenhall, Pratul~P Srinivasan, Matthew Tancik, Jonathan~T Barron, Ravi Ramamoorthi, and Ren Ng.
\newblock Nerf: Representing scenes as neural radiance fields for view synthesis.
\newblock \emph{Communications of the ACM}, 65\penalty0 (1):\penalty0 99--106, 2021.

\bibitem[Mirowski et~al.(2022)Mirowski, Pascanu, Viola, Soyer, Ballard, Banino, Denil, Goroshin, Sifre, Kavukcuoglu, et~al.]{mirowski2022learning}
Piotr Mirowski, Razvan Pascanu, Fabio Viola, Hubert Soyer, Andy Ballard, Andrea Banino, Misha Denil, Ross Goroshin, Laurent Sifre, Koray Kavukcuoglu, et~al.
\newblock Learning to navigate in complex environments.
\newblock In \emph{International Conference on Learning Representations}, 2022.

\bibitem[Nichol and Dhariwal(2021)]{pmlr-v139-nichol21a}
Alexander~Quinn Nichol and Prafulla Dhariwal.
\newblock Improved denoising diffusion probabilistic models.
\newblock In \emph{Proceedings of the 38th International Conference on Machine Learning}, pages 8162--8171. PMLR, 2021.

\bibitem[Pathak et~al.(2018)Pathak, Mahmoudieh, Luo, Agrawal, Chen, Shentu, Shelhamer, Malik, Efros, and Darrell]{pathak2018zero}
Deepak Pathak, Parsa Mahmoudieh, Guanghao Luo, Pulkit Agrawal, Dian Chen, Yide Shentu, Evan Shelhamer, Jitendra Malik, Alexei~A Efros, and Trevor Darrell.
\newblock Zero-shot visual imitation.
\newblock In \emph{Proceedings of the IEEE conference on computer vision and pattern recognition workshops}, pages 2050--2053, 2018.

\bibitem[Peebles and Xie(2023)]{Peebles_2023_ICCV}
William Peebles and Saining Xie.
\newblock Scalable diffusion models with transformers.
\newblock In \emph{Proceedings of the IEEE/CVF International Conference on Computer Vision (ICCV)}, pages 4195--4205, 2023.

\bibitem[Polyak et~al.(2024)Polyak, Zohar, Brown, Tjandra, Sinha, Lee, Vyas, Shi, Ma, Chuang, et~al.]{polyak2024movie}
Adam Polyak, Amit Zohar, Andrew Brown, Andros Tjandra, Animesh Sinha, Ann Lee, Apoorv Vyas, Bowen Shi, Chih-Yao Ma, Ching-Yao Chuang, et~al.
\newblock Movie gen: A cast of media foundation models.
\newblock \emph{arXiv preprint arXiv:2410.13720}, 2024.

\bibitem[Poole et~al.()Poole, Jain, Barron, and Mildenhall]{pooledreamfusion}
Ben Poole, Ajay Jain, Jonathan~T Barron, and Ben Mildenhall.
\newblock Dreamfusion: Text-to-3d using 2d diffusion.
\newblock In \emph{The Eleventh International Conference on Learning Representations}.

\bibitem[Ronneberger et~al.(2015)Ronneberger, Fischer, and Brox]{ronneberger2015u}
Olaf Ronneberger, Philipp Fischer, and Thomas Brox.
\newblock U-net: Convolutional networks for biomedical image segmentation.
\newblock In \emph{Medical image computing and computer-assisted intervention--MICCAI 2015: 18th international conference, Munich, Germany, October 5-9, 2015, proceedings, part III 18}, pages 234--241. Springer, 2015.

\bibitem[Rubinstein(1997)]{rubinstein1997optimization}
Reuven~Y Rubinstein.
\newblock Optimization of computer simulation models with rare events.
\newblock \emph{European Journal of Operational Research}, 99\penalty0 (1):\penalty0 89--112, 1997.

\bibitem[Savva et~al.(2019)Savva, Kadian, Maksymets, Zhao, Wijmans, Jain, Straub, Liu, Koltun, Malik, et~al.]{savva2019habitat}
Manolis Savva, Abhishek Kadian, Oleksandr Maksymets, Yili Zhao, Erik Wijmans, Bhavana Jain, Julian Straub, Jia Liu, Vladlen Koltun, Jitendra Malik, et~al.
\newblock Habitat: A platform for embodied ai research.
\newblock In \emph{Proceedings of the IEEE/CVF international conference on computer vision}, pages 9339--9347, 2019.

\bibitem[Seo et~al.(2023)Seo, Hafner, Liu, Liu, James, Lee, and Abbeel]{seo2023masked}
Younggyo Seo, Danijar Hafner, Hao Liu, Fangchen Liu, Stephen James, Kimin Lee, and Pieter Abbeel.
\newblock Masked world models for visual control.
\newblock In \emph{Conference on Robot Learning}, pages 1332--1344. PMLR, 2023.

\bibitem[Shah et~al.()Shah, Sridhar, Dashora, Stachowicz, Black, Hirose, and Levine]{shahvint}
Dhruv Shah, Ajay Sridhar, Nitish Dashora, Kyle Stachowicz, Kevin Black, Noriaki Hirose, and Sergey Levine.
\newblock Vint: A foundation model for visual navigation.
\newblock In \emph{7th Annual Conference on Robot Learning}.

\bibitem[Shah et~al.(2021)Shah, Eysenbach, Kahn, Rhinehart, and Levine]{shah2021rapid}
Dhruv Shah, Benjamin Eysenbach, Gregory Kahn, Nicholas Rhinehart, and Sergey Levine.
\newblock Rapid exploration for open-world navigation with latent goal models.
\newblock \emph{arXiv preprint arXiv:2104.05859}, 2021.

\bibitem[Shah et~al.(2023)Shah, Sridhar, Bhorkar, Hirose, and Levine]{shah2023gnm}
Dhruv Shah, Ajay Sridhar, Arjun Bhorkar, Noriaki Hirose, and Sergey Levine.
\newblock Gnm: A general navigation model to drive any robot.
\newblock In \emph{2023 IEEE International Conference on Robotics and Automation (ICRA)}, pages 7226--7233. IEEE, 2023.

\bibitem[Sohl-Dickstein et~al.(2015)Sohl-Dickstein, Weiss, Maheswaranathan, and Ganguli]{sohl2015deep}
Jascha Sohl-Dickstein, Eric Weiss, Niru Maheswaranathan, and Surya Ganguli.
\newblock Deep unsupervised learning using nonequilibrium thermodynamics.
\newblock In \emph{International conference on machine learning}, pages 2256--2265. PMLR, 2015.

\bibitem[Sridhar et~al.(2024)Sridhar, Shah, Glossop, and Levine]{sridhar2024nomad}
Ajay Sridhar, Dhruv Shah, Catherine Glossop, and Sergey Levine.
\newblock Nomad: Goal masked diffusion policies for navigation and exploration.
\newblock In \emph{2024 IEEE International Conference on Robotics and Automation (ICRA)}, pages 63--70. IEEE, 2024.

\bibitem[Srivastava et~al.(2017)Srivastava, Valkov, Russell, Gutmann, and Sutton]{srivastava2017veegan}
Akash Srivastava, Lazar Valkov, Chris Russell, Michael~U Gutmann, and Charles Sutton.
\newblock Veegan: Reducing mode collapse in gans using implicit variational learning.
\newblock \emph{Advances in neural information processing systems}, 30, 2017.

\bibitem[Sturm et~al.(2012)Sturm, Burgard, and Cremers]{sturm2012evaluating}
J{\"u}rgen Sturm, Wolfram Burgard, and Daniel Cremers.
\newblock Evaluating egomotion and structure-from-motion approaches using the tum rgb-d benchmark.
\newblock In \emph{Proc. of the Workshop on Color-Depth Camera Fusion in Robotics at the IEEE/RJS International Conference on Intelligent Robot Systems (IROS)}, page~6, 2012.

\bibitem[Thanh-Tung and Tran(2020)]{thanh2020catastrophic}
Hoang Thanh-Tung and Truyen Tran.
\newblock Catastrophic forgetting and mode collapse in gans.
\newblock In \emph{2020 international joint conference on neural networks (ijcnn)}, pages 1--10. IEEE, 2020.

\bibitem[Tomar et~al.(2024)Tomar, Hansen-Estruch, Bachman, Lamb, Langford, Taylor, and Levine]{tomar2024videooccupancymodels}
Manan Tomar, Philippe Hansen-Estruch, Philip Bachman, Alex Lamb, John Langford, Matthew~E. Taylor, and Sergey Levine.
\newblock Video occupancy models, 2024.

\bibitem[Triest et~al.(2022)Triest, Sivaprakasam, Wang, Wang, Johnson, and Scherer]{triest2022tartandrive}
Samuel Triest, Matthew Sivaprakasam, Sean~J Wang, Wenshan Wang, Aaron~M Johnson, and Sebastian Scherer.
\newblock Tartandrive: A large-scale dataset for learning off-road dynamics models.
\newblock In \emph{2022 International Conference on Robotics and Automation (ICRA)}, pages 2546--2552. IEEE, 2022.

\bibitem[Tulyakov et~al.(2018{\natexlab{a}})Tulyakov, Liu, Yang, and Kautz]{Tulyakov:2018:MoCoGAN}
Sergey Tulyakov, Ming-Yu Liu, Xiaodong Yang, and Jan Kautz.
\newblock {MoCoGAN}: Decomposing motion and content for video generation.
\newblock In \emph{IEEE Conference on Computer Vision and Pattern Recognition (CVPR)}, pages 1526--1535, 2018{\natexlab{a}}.

\bibitem[Tulyakov et~al.(2018{\natexlab{b}})Tulyakov, Liu, Yang, and Kautz]{tulyakov2018mocogan}
Sergey Tulyakov, Ming-Yu Liu, Xiaodong Yang, and Jan Kautz.
\newblock Mocogan: Decomposing motion and content for video generation.
\newblock In \emph{Proceedings of the IEEE conference on computer vision and pattern recognition}, pages 1526--1535, 2018{\natexlab{b}}.

\bibitem[Tung et~al.(2025)Tung, Chou, Cai, Yang, Zhang, Wetzstein, Hariharan, and Snavely]{cho}
Joseph Tung, Gene Chou, Ruojin Cai, Guandao Yang, Kai Zhang, Gordon Wetzstein, Bharath Hariharan, and Noah Snavely.
\newblock Megascenes: Scene-level view synthesis at scale.
\newblock In \emph{Computer Vision -- ECCV 2024}, pages 197--214, Cham, 2025. Springer Nature Switzerland.

\bibitem[Unterthiner et~al.(2019)Unterthiner, van Steenkiste, Kurach, Marinier, Michalski, and Gelly]{unterthiner2019fvd}
Thomas Unterthiner, Sjoerd van Steenkiste, Karol Kurach, Rapha{\"e}l Marinier, Marcin Michalski, and Sylvain Gelly.
\newblock Fvd: A new metric for video generation.
\newblock 2019.

\bibitem[Uria et~al.(2022)Uria, Ibarz, Banino, Zambaldi, Kumaran, Hassabis, Barry, and Blundell]{Uria2020.11.11.378141}
Benigno Uria, Borja Ibarz, Andrea Banino, Vinicius Zambaldi, Dharshan Kumaran, Demis Hassabis, Caswell Barry, and Charles Blundell.
\newblock A model of egocentric to allocentric understanding in mammalian brains.
\newblock \emph{bioRxiv}, 2022.

\bibitem[Valevski et~al.(2024)Valevski, Leviathan, Arar, and Fruchter]{valevski2024diffusion}
Dani Valevski, Yaniv Leviathan, Moab Arar, and Shlomi Fruchter.
\newblock Diffusion models are real-time game engines.
\newblock \emph{arXiv preprint arXiv:2408.14837}, 2024.

\bibitem[Van~Hoorick et~al.(2024)Van~Hoorick, Wu, Ozguroglu, Sargent, Liu, Tokmakov, Dave, Zheng, and Vondrick]{vanhoorick2024gcd}
Basile Van~Hoorick, Rundi Wu, Ege Ozguroglu, Kyle Sargent, Ruoshi Liu, Pavel Tokmakov, Achal Dave, Changxi Zheng, and Carl Vondrick.
\newblock Generative camera dolly: Extreme monocular dynamic novel view synthesis.
\newblock 2024.

\bibitem[Vaswani(2017)]{vaswani2017attention}
A Vaswani.
\newblock Attention is all you need.
\newblock \emph{Advances in Neural Information Processing Systems}, 2017.

\bibitem[Voleti et~al.(2022)Voleti, Jolicoeur-Martineau, and Pal]{voleti2022mcvd}
Vikram Voleti, Alexia Jolicoeur-Martineau, and Chris Pal.
\newblock Mcvd-masked conditional video diffusion for prediction, generation, and interpolation.
\newblock \emph{Advances in neural information processing systems}, 35:\penalty0 23371--23385, 2022.

\bibitem[Wang et~al.(2024)Wang, Huang, Bergman, Shen, Gao, Lingelbach, Sun, Bian, Song, Liu, et~al.]{wang2024phased}
Fu-Yun Wang, Zhaoyang Huang, Alexander Bergman, Dazhong Shen, Peng Gao, Michael Lingelbach, Keqiang Sun, Weikang Bian, Guanglu Song, Yu Liu, et~al.
\newblock Phased consistency models.
\newblock \emph{Advances in Neural Information Processing Systems}, 37:\penalty0 83951--84009, 2024.

\bibitem[Wu et~al.(2023)Wu, Escontrela, Hafner, Abbeel, and Goldberg]{wu2023daydreamer}
Philipp Wu, Alejandro Escontrela, Danijar Hafner, Pieter Abbeel, and Ken Goldberg.
\newblock Daydreamer: World models for physical robot learning.
\newblock In \emph{Conference on robot learning}, pages 2226--2240. PMLR, 2023.

\bibitem[Xu et~al.(2019)Xu, Sun, Zhang, Zhao, and Lin]{xu2019understandingimprovinglayernormalization}
Jingjing Xu, Xu Sun, Zhiyuan Zhang, Guangxiang Zhao, and Junyang Lin.
\newblock Understanding and improving layer normalization, 2019.

\bibitem[Yang et~al.()Yang, Du, Ghasemipour, Tompson, Kaelbling, Schuurmans, and Abbeel]{yanglearning}
Sherry Yang, Yilun Du, Seyed Kamyar~Seyed Ghasemipour, Jonathan Tompson, Leslie~Pack Kaelbling, Dale Schuurmans, and Pieter Abbeel.
\newblock Learning interactive real-world simulators.
\newblock In \emph{The Twelfth International Conference on Learning Representations}.

\bibitem[Yu et~al.(2023)Yu, Cheng, Sohn, Lezama, Zhang, Chang, Hauptmann, Yang, Hao, Essa, et~al.]{yu2023magvit}
Lijun Yu, Yong Cheng, Kihyuk Sohn, Jos{\'e} Lezama, Han Zhang, Huiwen Chang, Alexander~G Hauptmann, Ming-Hsuan Yang, Yuan Hao, Irfan Essa, et~al.
\newblock Magvit: Masked generative video transformer.
\newblock In \emph{Proceedings of the IEEE/CVF Conference on Computer Vision and Pattern Recognition}, pages 10459--10469, 2023.

\bibitem[Zhang et~al.(2018{\natexlab{a}})Zhang, Isola, Efros, Shechtman, and Wang]{zhang2018perceptual}
Richard Zhang, Phillip Isola, Alexei~A Efros, Eli Shechtman, and Oliver Wang.
\newblock The unreasonable effectiveness of deep features as a perceptual metric.
\newblock In \emph{CVPR}, 2018{\natexlab{a}}.

\bibitem[Zhang et~al.(2018{\natexlab{b}})Zhang, Isola, Efros, Shechtman, and Wang]{zhang2018unreasonable}
Richard Zhang, Phillip Isola, Alexei~A Efros, Eli Shechtman, and Oliver Wang.
\newblock The unreasonable effectiveness of deep features as a perceptual metric.
\newblock In \emph{Proceedings of the IEEE conference on computer vision and pattern recognition}, pages 586--595, 2018{\natexlab{b}}.

\bibitem[Zhou et~al.(2024)Zhou, Pan, LeCun, and Pinto]{zhou2024dinowmworldmodelspretrained}
Gaoyue Zhou, Hengkai Pan, Yann LeCun, and Lerrel Pinto.
\newblock Dino-wm: World models on pre-trained visual features enable zero-shot planning, 2024.

\end{thebibliography}
}
\clearpage
\setcounter{page}{1}
\maketitlesupplementary
\label{sec:supp}

The structure of the Appendix is as follows: we start by describing how we plan navigation trajectories via Standalone Planning in Section~\ref{sec:supp:optimization}, and then include more experiments and results in Section~\ref{supp:sec:exp}.

\section{Standalone Planning Optimization}
\label{sec:supp:optimization}

As described in Section~\ref{sec:methods-navi}, we use a pretrained NWM to standalone-plan goal-conditioned navigation trajectories by optimizing Eq.\ref{eq:planning-loss}. Here, we provide additional details about the optimization using the Cross-Entropy Method~\citep{rubinstein1997optimization} and the hyperparameters used. Full standalone navigation planning results are presented in Section~\ref{sec:supp:results}.

We optimize trajectories using the Cross-Entropy Method, a gradient-free stochastic optimization technique for continuous optimization problems. This method iteratively updates a probability distribution to improve the likelihood of generating better solutions. In the unconstrained standalone planning scenario, we assume the trajectory is a straight line and optimize only its endpoint, represented by three variables: a single translation $u$ and yaw rotation $\phi$. We then map this tuple into eight evenly spaced delta steps, applying the yaw rotation at the final step. The time interval between steps is fixed at $k=0.25$ seconds. The main steps of our optimization process are as follows:

\begin{itemize} 
    \item \textbf{Initialization}: Define a Gaussian distribution with mean $\mu = (\mu_{\Delta x}, \mu_{\Delta y}, \mu_\phi)$ and variance $\Sigma = \mathrm{diag}(\sigma_{\Delta x}^2, \sigma_{\Delta y}^2, \sigma_\phi^2)$ over the solution space.
    
    \item \textbf{Sampling}: Generate $N=120$ candidate solutions by sampling from the current Gaussian distribution.
    
    \item \textbf{Evaluation}: Evaluate each candidate solution by simulating it using the NWM and measuring the LPIPS score between the simulation output and input goal images. Since NWM is stochastic, we evaluate each candidate solution $M$ times and average to obtain a final score.
    
    \item \textbf{Selection}: Select a subset of the best-performing solutions based on the LPIPS scores.
    
    \item \textbf{Update}: Adjust the parameters of the distribution to increase the probability of generating solutions similar to the top-performing ones. This step minimizes the cross-entropy between the old and updated distributions.
    
    \item \textbf{Iteration}: Repeat the sampling, evaluation, selection, and update steps until a stopping criterion (e.g. convergence or iteration limit) is met.
\end{itemize}
For simplicity, we run the optimization process for a single iteration, which we found effective for short-horizon planning of two seconds, though further improvements are possible with more iterations. When navigation constraints are applied, parts of the trajectory are zeroed out to respect these constraints. For instance, in the "forward-first" scenario, the translation action is $u=(\Delta x, 0)$ for the first five steps and $u=(0, \Delta y)$ for the last three steps.


\begin{table*}[th!]
\small
\centering
\begin{tabular}{l|cccccccccc}
\hline
& \multicolumn{1}{c}{unknown environment} & \multicolumn{4}{c}{known environments}\\
data & Go Stanford & RECON & HuRoN & SCAND & TartanDrive \\

\hline
in-domain data & $0.658 \pm 0.002$ & $\textbf{0.295} \pm 0.002$ & $\textbf{0.250} \pm 0.003$ & $0.403 \pm 0.002$ & $\textbf{0.414} \pm 0.001$ \\
+ Ego4D~\text{\footnotesize(unlabeled)} & $\textbf{0.652} \pm 0.003$ & $0.368 \pm 0.003$ & $0.377 \pm 0.002$ & $\textbf{0.398} \pm 0.001$ & $0.430 \pm 0.000$ \\ 
\hline
\end{tabular}
\caption{\textbf{Training on additional unlabeled data improves performance on  unseen environments. } Reporting results on unknown environment (Go Stanford) and known one (RECON). Results reported by evaluating LPIPS $4$ seconds into the future.}
\label{tab:sup:oodm}
\end{table*}

\section{Experiments and Results}
\label{supp:sec:exp}
\subsection{Experimental Study}
\label{supp:sec:exp-study}
We elaborate on the metrics and datasets used.

\noindent\textbf{Evaluation Metrics.} 
We describe the evaluation metrics used to assess predicted navigation trajectories and the quality of images generated by our NWM.

For visual navigation performance, \textbf{Absolute Trajectory Error} (ATE) measures the overall accuracy of trajectory estimation by computing the Euclidean distance between corresponding points in the estimated and ground-truth trajectories. \textbf{Relative Pose Error} (RPE) evaluates the consistency of consecutive poses by calculating the error in relative transformations between them~\citep{sturm2012evaluating}.

To more rigorously assess the semantics in the world model outputs, we use {Learned Perceptual Image Patch Similarity} (LPIPS) and {DreamSim}~\citep{fu2024dreamsim}, which evaluate perceptual similarity by comparing deep features from a neural network~\citep{zhang2018perceptual}. LPIPS, in particular, uses AlexNet~\citep{krizhevsky2012imagenet} to focus on human perception of structural differences. Additionally, we use {Peak Signal-to-Noise Ratio} (PSNR) to quantify the pixel-level quality of generated images by measuring the ratio of maximum pixel value to error, with higher values indicating better quality.

To study image and video synthesis quality, we use {Fréchet Inception Distance} (FID) and {Fréchet Video Distance} (FVD), which compare the feature distributions of real and generated images or videos. Lower FID and FVD scores indicate higher visual quality~\citep{heusel2017gans, unterthiner2019fvd}.

\vspace{1.3mm}
\noindent\textbf{Datasets}. For all robotics datasets, we have access to the location and rotation of the robots, and we use this to infer the actions as the delta in location and rotation. We remove all backward movement which can be jittery following NoMaD~\citep{sridhar2024nomad}, thereby splitting the data to forward walking segments for SCAND~\citep{karnan2022socially}, TartanDrive~\citep{triest2022tartandrive}, RECON~\citep{shah2021rapid}, and HuRoN~\citep{hirose2023sacson}. We also utilize unlabeled Ego4D videos, where we only use time shift as action. Next, we describe each individual dataset.

\begin{itemize}
    \item SCAND~\citep{karnan2022socially} is a robotics dataset consisting of socially compliant navigation demonstrations using a wheeled Clearpath Jackal and a legged Boston Dynamics Spot. SCAND has demonstrations in both indoor and outdoor settings at UT Austin. The dataset consists of $8.7$ hours, $138$ trajectories, $25$ miles of data and we use the corresponding camera poses. We use $484$ video segments for training and $121$ video segments for testing. Used for training and evaluation.
    \item TartanDrive~\citep{triest2022tartandrive} is an outdoor off-roading driving dataset collected using a modified Yamaha Viking ATV in Pittsburgh. The dataset consists of $5$ hours and $630$ trajectories. We use $1,000$ video segments for training and $251$ video segments for testing. 
    \item RECON~\citep{shah2021rapid} is an outdoor robotics dataset collected using a Clearpath Jackal UGV platform. The dataset consists of $40$ hours across $9$ open-world environments. We use $9,468$ video segments for training and $2,367$ video segments for testing. Used for training and evaluation.
    \item HuRoN~\citep{hirose2023sacson} is a robotics dataset consisting of social interactions using a Robot Roomba in indoor settings collected at UC Berkeley. The dataset consists of over $75$ hours in $5$ different environments with $4,000$ human interactions. We use $2,451$ video segments for training and $613$ video segments for testing. Used for training and evaluation.
    \item GO Stanford~\citep{hirose2018gonet,hirose2019vunet}, a robotics datasets capturing the fisheye video footage of two different teleoperated robots, collected at at least $27$ different Stanford building with around $25$ hours of video footage. Due to the low resolution images, we only use it for out of domain evaluation.

    \item Ego4D~\citep{grauman2022ego4d} is a large-scale egocentric dataset consisting of $3,670$ hours across $74$ locations. Ego4D consists a variety of scenarios such as Arts $\&$ Crafts, Cooking, Construction, Cleaning $\&$ Laundry, and Grocery Shopping. We use only use videos which involve visual navigation such as Grocery Shopping and Jogging. We use a total $1619$ videos of over $908$ hours for training only. Only used for unlabeled training unlabeled training. The videos we use are from the following Ego4D scenarios: ``Skateboard/scooter'', ``Roller skating'', ``Football'', ``Attending a festival or fair'', ``Gardener'', ``Mini golf'', ``Riding motorcycle'', ``Golfing'', ``Cycling/jogging'', ``Walking on street'', ``Walking the dog/pet'', ``Indoor Navigation (walking)'', ``Working in outdoor store'', ``Clothes/other shopping'', ``Playing with pets'', ``Grocery shopping indoors'', ``Working out outside'', ``Farmer'', ``Bike'', ``Flower Picking'', ``Attending sporting events (watching and participating)'', ``Drone flying'', ``Attending a lecture/class'', ``Hiking'', ``Basketball'', ``Gardening'', ``Snow sledding'', ``Going to the park''.
\end{itemize}

\noindent\textbf{Visual Navigation Evaluation Set.} Our main finding when constructing visual navigation evaluation sets is that forward motion is highly prevalent, and if not carefully accounted for, it can dominate the evaluation data. To create diverse evaluation sets, we rank potential evaluation trajectories based on how well they can be predicted by simply moving forward. For each dataset, we select the $100$ examples that are least predictable by this heuristic and use them for evaluation.

\vspace{1.3mm} \noindent\textbf{Time Prediction Evaluation Set.} Predicting the future frame after $k$ seconds is more challenging than estimating a trajectory, as it requires both predicting the agent's trajectory and its orientation in pixel space. Therefore, we do not impose additional diversity constraints. For each dataset, we randomly select $500$ test prediction examples.

\subsection{Experiments and Results} 
\label{sec:supp:results}

\begin{figure*}
    \centering
\includegraphics[width=1\linewidth]{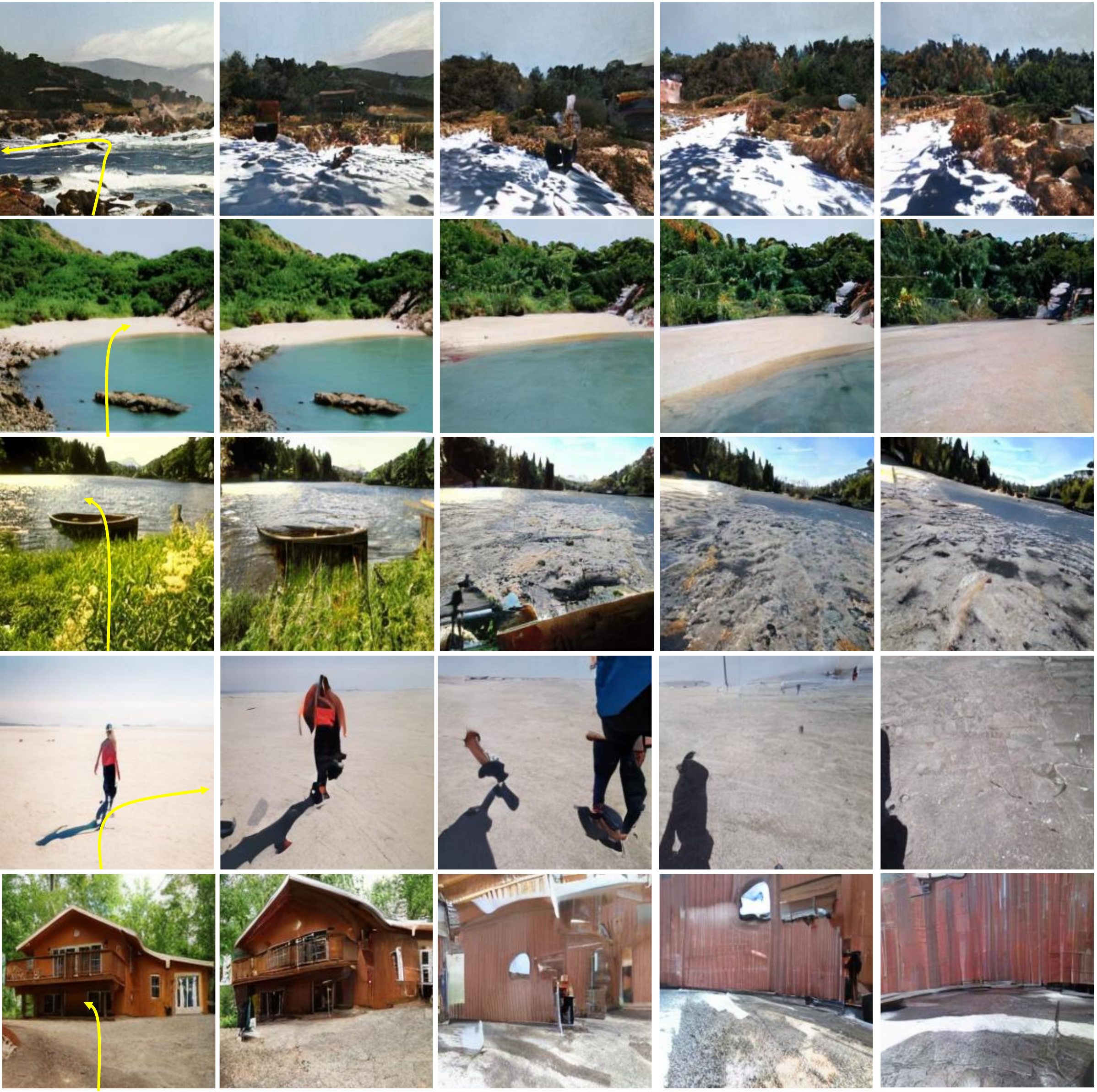}
    \caption{\textbf{Navigating Unknown Environments}. NWM is conditioned on a single image, and autoregressively predicts the next states given the associated actions (marked in yellow) up to $4$ seconds and $4$ FPS. We plot the generated results after 1, 2, 3, and 4 seconds.}
    \label{fig:supp:unkown}
\end{figure*}

\noindent\textbf{Training on Additional Unlabeled Data.} We include results for additional known environments in Table~\ref{tab:sup:oodm} and Figure~\ref{fig:supp:unkown}. We find that in known environments, models trained exclusively with in-domain data tend to perform better, likely because they are better tailored to the in-domain distribution. The only exception is the SCAND dataset, where dynamic objects (e.g. humans walking) are present. In this case, adding unlabeled data may help improve performance by providing additional diverse examples.

\begin{figure*}
    \centering
\includegraphics[width=0.94\linewidth]{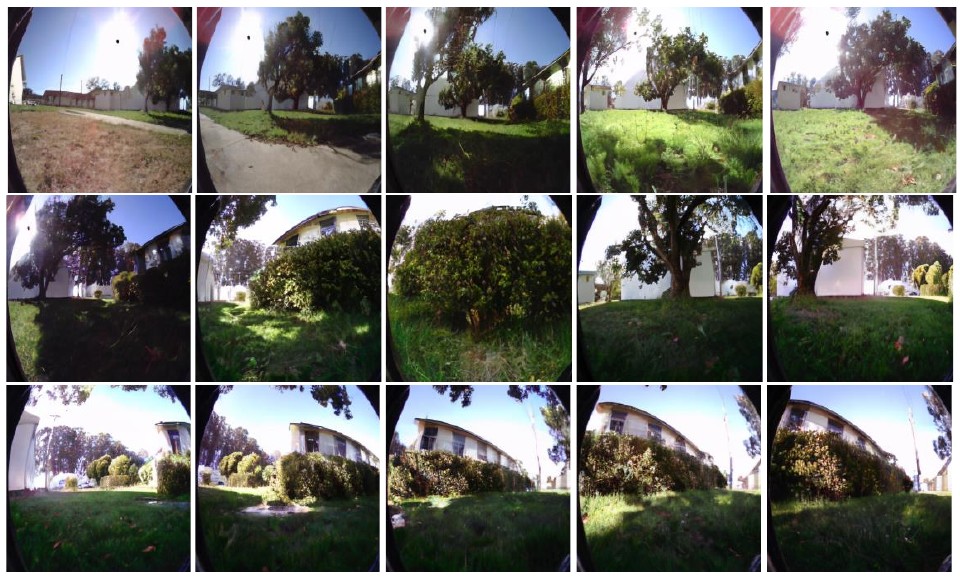}
    \caption{\textbf{Video generation examples on RECON}. NWM is conditioned on a single first image, and a ground truth trajectory and autoregressively predicts the next up to $16$ seconds at $4$ FPS. We plot the generated results from $2$ to $16$ seconds, every 1 second.}
    \label{fig:supp:known_recon}
\end{figure*}

\begin{figure*}
    \centering
\includegraphics[width=0.94\linewidth]{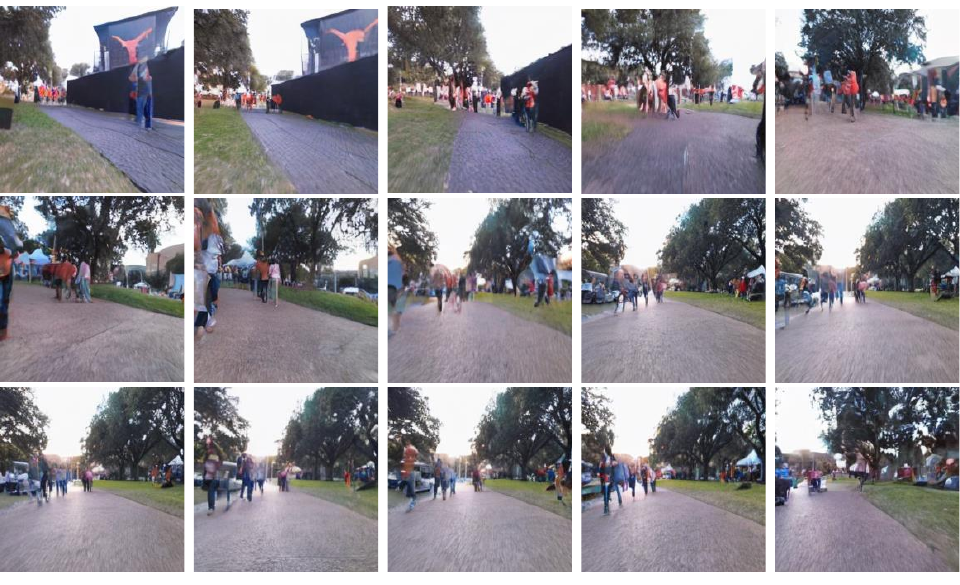}
    \caption{\textbf{Video generation examples on SCAND}. NWM is conditioned on a single first image, and a ground truth trajectory and autoregressively predicts the next up to $16$ seconds at $4$ FPS. We plot the generated results from $2$ to $16$ seconds, every 1 second.}
    \label{fig:supp:known_scand}
\end{figure*}

\begin{figure*}
    \centering
\includegraphics[width=0.94\linewidth]{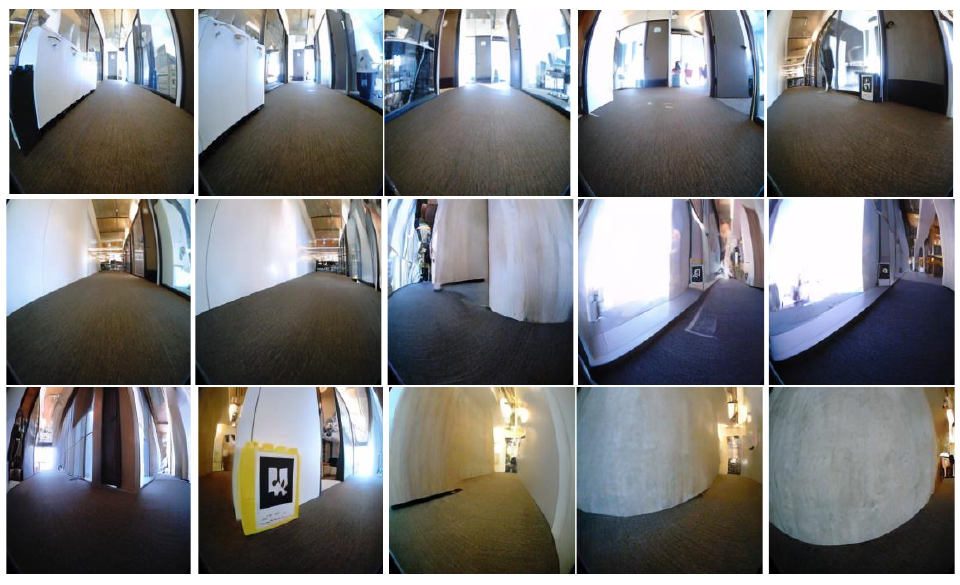}
    \caption{\textbf{Video generation examples on HuRoN}. NWM is conditioned on a single first image, and a ground truth trajectory and autoregressively predicts the next up to $16$ seconds at $4$ FPS. We plot the generated results from $2$ to $16$ seconds, every 1 second.}
    \label{fig:supp:known_huron}
\end{figure*}

\begin{figure*}
    \centering
\includegraphics[width=0.94\linewidth]{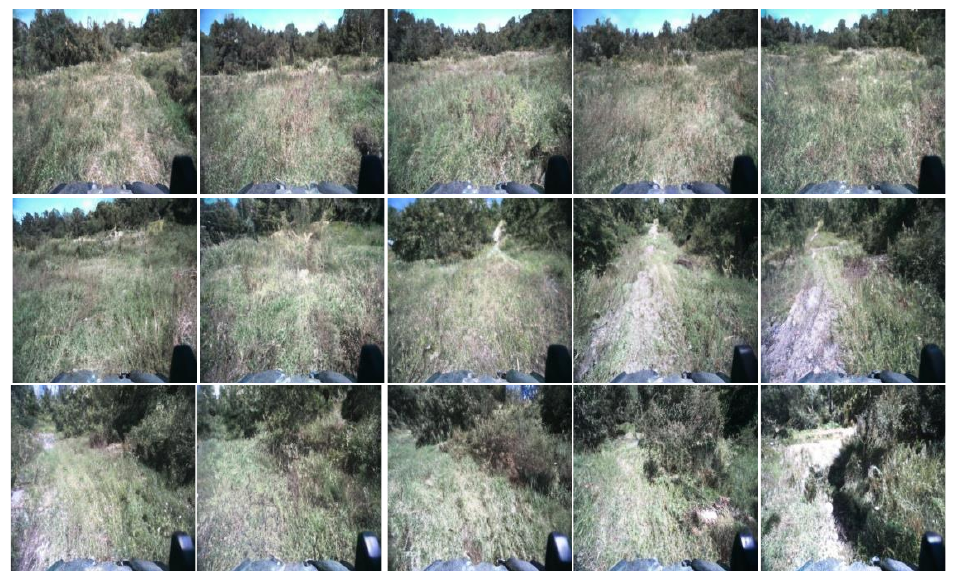}
    \caption{\textbf{Video generation examples on Tartan Drive}. NWM is conditioned on a single first image, and a ground truth trajectory and autoregressively predicts the next up to $16$ seconds at $4$ FPS. We plot the generated results from $2$ to $16$ seconds, every 1 second.}
    \label{fig:supp:known_tartan}
\end{figure*}

\noindent\textbf{Known Environments.} We include additional visualization results of following trajectories using NWM in the known environments RECON (Figure~\ref{fig:supp:known_recon}), SCAND (Figure~\ref{fig:supp:known_scand}), HuRoN (Figure~\ref{fig:supp:known_huron}), and Tartan Drive (Figure~\ref{fig:supp:known_tartan}). Additionally, we include full FVD comparison of DIAMOND and NWM in Table~\ref{tab:supp:fvd}.
\begin{table}[H]
\small
\centering
\begin{tabular}{l|ccccc}
\hline
dataset & DIAMOND & NWM (ours) \\
\hline
RECON & $762.734 \pm 3.361$ & $\textbf{200.969} \pm 5.629$ \\
HuRoN & $881.981 \pm 11.601$ & $\textbf{276.932} \pm 4.346$ \\
TartanDrive & $2289.687 \pm 6.991$ & $\textbf{494.247} \pm 14.433$ \\
SCAND & $1945.085 \pm 8.449$ & $\textbf{401.699} \pm 11.216$ \\
\hline
\end{tabular}
\captionof{table}{\textbf{Comparison of Video Synthesis Quality.} $16$ second videos generated at 4 FPS, reporting FVD (lower is better).}
\label{tab:supp:fvd}
\end{table}

\begin{table*}[t!]
\footnotesize
\centering
\resizebox{\textwidth}{!}{%
\begin{tabular}{l|cccccccccccc}
\hline
model & \multicolumn{2}{c}{RECON} & \multicolumn{2}{c}{HuRoN} & \multicolumn{2}{c}{Tartan} & \multicolumn{2}{c}{SCAND}\\
& ATE & RTE & ATE & RTE & ATE & RTE & ATE & RTE\\
\hline
Forward & $1.92$ $\pm$ $0.00$ & $0.54$ $\pm$ $0.00$ & $4.14$ $\pm$ $0.00$ & $1.05$ $\pm$ $0.00$ & $5.75$ $\pm$ $0.00$ & $1.19$ $\pm$ $0.00$ & $2.97$ $\pm$ $0.00$ & $0.62$ $\pm$ $0.00$ \\
GNM &  $1.87$ $\pm$ $0.00$ & $0.73$ $\pm$ $0.00$ & $3.71$ $\pm$ $0.00$ & $1.00$ $\pm$ $0.00$ & $6.65$ $\pm$ $0.00$ & $1.62$ $\pm$ $0.00$ & $2.12$ $\pm$ $0.00$ & $0.61$ $\pm$ $0.00$ \\
NoMaD &  $1.95$ $\pm$ $0.05$ & $0.53$ $\pm$ $0.01$ & $3.73$ $\pm$ $0.04$ & $0.96$ $\pm$ $0.01$ & $6.32$ $\pm$ $0.03$ & $1.31$ $\pm$ $0.01$ & $2.24$ $\pm$ $0.03$ & $0.49$ $\pm$ $0.01$ \\
NWM + NoMaD ($\times 16$) &  $1.88$ $\pm$ $0.03$ & $0.51$ $\pm$ $0.01$ & $3.73$ $\pm$ $0.05$ & $0.95$ $\pm$ $0.01$ & $6.26$ $\pm$ $0.06$ & $1.30$ $\pm$ $0.01$ & $2.18$ $\pm$ $0.05$ & $0.48$ $\pm$ $0.01$ \\
NWM + NoMaD ($\times 32$) & 1.79 $\pm$ $0.02$ & $0.49$ $\pm$ $0.00$ & $\textbf{3.68}$ $\pm$ $0.03$ & $\textbf{0.95}$ $\pm$ $0.01$ & $6.25$ $\pm$ $0.05$ & $1.29$ $\pm$ $0.01$ & $2.19$ $\pm$ $0.03$ & $0.47$ $\pm$ $0.01$ \\
\hline
NWM (only) &  $\textbf{1.13}$ $\pm$ $0.02$ & $\textbf{0.35}$ $\pm$ $0.01$  & $4.12$ $\pm$ $0.03$ & $0.96$ $\pm$ $0.01$ & $\textbf{5.63}$ $\pm$ $0.06$ & $\textbf{1.18}$ $\pm$ $0.01$ & $\textbf{1.28}$ $\pm$ $0.02$ & $\textbf{0.33}$ $\pm$ $0.01$\\
\hline
\end{tabular}%
}
\caption{\textbf{Goal Conditioned Visual Navigation}. ATE and RPE results on on all in domain datasets, predicting trajectories of up to $2$ seconds. NWM achieves improved results on all metrics compared to previous approaches NoMaD~\citep{sridhar2024nomad} and GNM~\citep{shah2023gnm}.}
\label{tab:supp:navigation}
\end{table*}
\noindent\textbf{Planning (Ranking).} Full goal-conditioned navigation results for all in-domain datasets are presented in Table~\ref{tab:supp:navigation}. Compared to NoMaD, we observe consistent improvements when using NWM to select from a pool of $16$ trajectories, with further gains when selecting from a larger pool of $32$. For Tartan Drive, we note that the dataset is heavily dominated by forward motion, as reflected in the results compared to the "Forward" baseline, a prediction model that always selects forward-only motion.

\vspace{3mm}
\noindent\textbf{Standalone Planning.} For standalone planning, we run the optimization procedure outlined in Section~\ref{sec:supp:optimization} for $1$ step, and evaluate each trajectories for 3 times. For all datasets, we initialize $\mu_{\Delta y}$ and $\mu_\phi$ to be 0, and $\sigma_{\Delta y}^2$ and $\sigma_\phi^2$ to be 0.1. We use different $(\mu_{\Delta x}, \sigma_{\Delta x}^2)$ across each dataset: $(-0.1, 0.02)$ for RECON, $(0.5, 0.07)$ for TartanDrive, $(-0.25, 0.04)$ for SCAND, and $(-0.33, 0.03)$ for HuRoN. We include the full standalone navigation planning results in Table~\ref{tab:supp:navigation}. We find that using planning in the stand-alone setting performs better compared to other approaches, and specifically previous hard-coded policies.

\vspace{3mm}
\noindent\textbf{Real-World Applicability}.
A key bottleneck in deploying NWM in real-world robotics is inference speed. We evaluate methods to improve NWM efficiency and measure their impact on runtime. We focus on using NWM with a generative policy (Section~\ref{sec:methods-navi}) to rank $32$ four-second trajectories. Since trajectory evaluation is parallelizable, we analyze the runtime of simulating a single trajectory. We find that existing solutions can already enable real-time applications of NWM at 2-10HZ (Table~\ref{tab:runtime}).

\begin{table}[h] 
\small \centering
\begin{tabular}{cccc} \hline NWM & +Time Skip & +Distillation. & +Quant. 4-bit \\ 
$30.3 \pm 0.2$ & $14.7 \pm 0.1$ & $0.4 \pm 0.1$ & $0.1$ \text{(est.~\cite{frantar2022gptq})}\\ 
\hline 
\end{tabular} 
\caption{Runtime (seconds) on an NVIDIA RTX 6000 Ada card.}
\label{tab:runtime} 
\end{table}

Inference time can be accelerated by composing every adjacent pair of actions (via Eq.~\ref{eq:compose-actions}) then simulating only $8$ future states instead of $16$ (``Time Skip''), which does not degrade navigation performance. Reducing the diffusion denoising steps from $250$ to $6$ by model distillation~\cite{wang2024phased} further speeds up inference with minor visual quality loss.\footnote{Using the distillation implementation for DiTs from~\url{https://github.com/hao-ai-lab/FastVideo}} Taken together, these two ideas can enable NWM to run in real time. Quantization to 4-bit, which we haven't explored, can lead to a $\times4$ speedup without performance hit~\cite{frantar2022gptq}.

\vspace{3mm}
\noindent\textbf{Test-time adaptation}. Test-time adaptation has shown to improve visual navigation~\cite{frey2023fast,pmlr-v235-gao24p}. What is the relation between planning using a world model and test-time adaptation? We hypothesize that the two ideas are orthogonal, and include test-time adaptation results. We consider a simplified adaptation approach by fine-tuning NWM for $2$k steps on trajectories from an \textit{unknown environment}. We show that this adaptation improves trajectory simulation in this environment (see ``ours+TTA'' in Table~\ref{tab:res}), where we also include additional baselines and ablations. 
\begin{table}[t] 
\footnotesize \centering
\begin{tabular}{cccccccc} \hline 
CDiT-L & context 2& action only & goals 2 & ours & ours + TTA \\ 
$0.656$&$0.655$&$0.661$&$0.654$&$0.652$ & $\textbf{0.650}$ \\ 
\hline 
\end{tabular} 
\caption{Results in \textit{unknown environment} (``Go Stanford''). Reporting lpips on $4$ seconds future prediction. Lower is better.} 
\label{tab:res} 
\end{table}


\end{document}